\documentclass[letterpaper]{article} 
\usepackage{aaai23}  
\usepackage{times}  
\usepackage{helvet}  
\usepackage{courier}  
\usepackage[hyphens]{url}  
\usepackage{graphicx} 
\usepackage{natbib}  
\usepackage{caption} 
\usepackage{algorithm}
\usepackage{algorithmic}

\usepackage{epsfig}
\usepackage{graphicx}
\usepackage{amsmath}

\usepackage{amssymb}
\usepackage{booktabs}
\usepackage{subfigure}
\usepackage{multirow}
\usepackage{color}
\usepackage{url}
\usepackage{newfloat}
\usepackage{listings}
\DeclareCaptionStyle{ruled}{labelfont=normalfont,labelsep=colon,strut=off} 
\floatstyle{ruled}
\newfloat{listing}{tb}{lst}{}
\floatname{listing}{Listing}
\title{CoordFill: Efficient High-Resolution Image Inpainting via \\ Parameterized Coordinate Querying}
\author {
    Weihuang Liu,\textsuperscript{\rm 1}
    Xiaodong Cun,\textsuperscript{\rm 2}$^{*}$
    Chi-Man Pun,\textsuperscript{\rm 1}$^{*}$
    Menghan Xia,\textsuperscript{\rm 2}
    Yong Zhang,\textsuperscript{\rm 2}
    Jue Wang\textsuperscript{\rm 2}
}
\affiliations {
    \textsuperscript{\rm 1} University of Macau,
    \textsuperscript{\rm 2} Tencent AI Lab\\
    mc05379@umac.mo, 
    vinthony@gmail.com, 
    cmpun@umac.mo,\\
    menghanxyz@gmail.com,
    zhangyong201303@gmail.com,
    arphid@gmail.com
}

\usepackage{bibentry}

\begin{document}

\twocolumn[{%
\renewcommand\twocolumn[1][]{#1}%
\maketitle
\begin{center}
    \centering
    \vspace{-2em}
    \includegraphics[width=\textwidth,height=5cm]{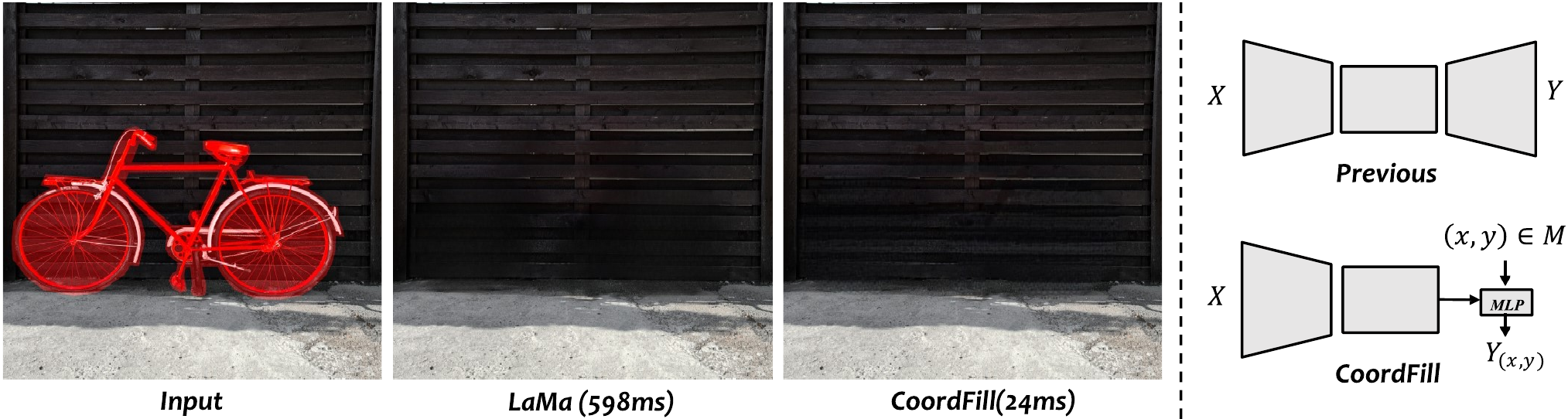}
    \vspace{-2em}
    \captionof{figure}{The proposed method only requires \underline{\textit{24ms}}~(25$\times$ faster than state-of-the-art method LaMa~\cite{lama} in the same resolution) to remove the red masked regions on the resolution of a 2048$\times$2048 image. Compared with most of the previous methods that synthesize the full image $Y$ by convolutional neural networks, CoordFill queries the coordinates $(x,y)$ in the missing mask $M$ of the input $X$ only and generates the pixel-value by the implicit representation. Thanks to the proposed decoder, CoordFill runs faster than previous methods on the high-resolution image inpainting task.  }
\end{center}%
}]



\newenvironment{alphafootnotes}
  {\par\edef\savedfootnotenumber{\number\value{footnote}}
  \renewcommand{\thefootnote}{\alph{footnote}}
  \setcounter{footnote}{0}}
  {\par\setcounter{footnote}{\savedfootnotenumber}}

\begin{alphafootnotes}
\let\thefootnote\relax\footnotetext{* Corresponding author}
\let\thefootnote\relax\footnotetext{Copyright \copyright\space 2023,
Association for the Advancement of Artificial Intelligence (www.aaai.org). All rights reserved.}
\end{alphafootnotes}

\begin{abstract}
Image inpainting aims to fill the missing hole of the input. It is hard to solve this task efficiently when facing high-resolution images due to two reasons: (1) Large reception field needs to be handled for high-resolution image inpainting. (2) The general encoder and decoder network synthesizes many background pixels synchronously due to the form of the image matrix. 
In this paper, we try to break the above limitations for the first time thanks to the recent development of continuous implicit representation. In detail, we down-sample and encode the degraded image to produce the spatial-adaptive parameters for each spatial patch via an attentional Fast Fourier Convolution~(FFC)-based parameter generation network. Then, we take these parameters as the weights and biases of a series of multi-layer perceptron~(MLP), where the input is the encoded continuous coordinates and the output is the synthesized color value. Thanks to the proposed structure, we only encode the high-resolution image in a relatively low resolution for larger reception field capturing. Then, the continuous position encoding will be helpful to synthesize the photo-realistic high-frequency textures by re-sampling the coordinate in a higher resolution. Also, our framework enables us to query the coordinates of missing pixels only in parallel, yielding a more efficient solution than the previous methods. Experiments show that the proposed method achieves real-time performance on 
the 2048$\times$2048 images using a single GTX 2080 Ti GPU and can handle 4096$\times$4096 images, with much better performance than existing state-of-the-art methods visually and numerically. The code is available at:~\url{https://github.com/NiFangBaAGe/CoordFill}.
\end{abstract}

\section{Introduction}

With the rapid development of digital media, the demand for image editing, for example, filling the given hole according to the remaining background pixels or removing the unnecessary object, also increases a lot. These tasks belong to image inpainting, which is one of the fundamental image synthesis tasks, which requires both semantic understanding and conditional generation.

Convolutional neural network~(CNN)-based methods dominate this field in recent years. For example, coarse-to-fine network structures and generative adversarial networks~\cite{gan} based methods~\cite{deepfill, deepfillv2, global_and_local, pconv} are used to synthesize the realistic textures. Other types of image inpainting are inspired by the typical hints~(\emph{e.g.} edges~\cite{edgeconnect}, semantic maps~\cite{semantic_guide}) from the original background, these methods guess the hints in the hole region firstly, and then, they synthesize the texture by the guidance of the filled hints. However, current methods still suffer from many limitations for real-world applications since casual images are in high resolution and on the mobile platform. We raise a question: \textit{What makes the inpainting hard in efficiently handling the high-resolution images?} and we attempt to answer this question from the following perspectives:

$(i)$ Learning the larger reception fields is time-consuming for high-resolution images. A larger reception field is essential in image inpainting~\cite{lama}, which means that we need to feed the full-size images to the inpainting network. In general, there are two straightforward ways to increase speed. We can crop or resize the images to fit the resolution of the input. However, direct cropping will influence the semantic guidance from the background region and hugely reduce the ability of scene understanding. Directly resizing is another option, it requires preserving the detailed high-frequency details for upsampling.

$(ii)$ The decoders will also synthesize a lot of unnecessary pixels. Image inpainting aims to remove a relatively small hole from the input. Restricted by the form of the image matrix, small irregular holes also need to be synthesized using a stack of convolutional layers in full image size.

We try to break the above limitations via the implicit representation from neural rendering community~\cite{siren,nerf,asapnet}. A typical implicit representation overfits the scene to small multi-layer perceptrons~(MLPs) network, where the input is the coordinate of the image matrix and the output is the queried pixels. It is a continuous representation that can keep the high-frequency details well and can achieve pixel-wise reconstruction. Inspired by these wonderful features, we propose a novel framework for efficient high-resolution image inpainting using implicit representation for the first time. To make the network related to the specific input, our method utilize a meta-learning strategy where the parameters of the pixel-wise query network are spatial-adaptive~\cite{asapnet}. In detail, our framework contains two sub-networks for parameter generation and pixel-wise query. 
In the parameter generation network, we resize the high-resolution image and generate the parameters for each local image patch. This network is built via a series of the proposed attentional FFC blocks. 
Then, the pixel-wise query network reforms the predicted parameters as a series of the MLP, where the input is the positional embeddings in the frequency domain and the output is the predicted color value in the given coordinate. Thanks to the proposed structure, our method can run faster in the challenge of high-resolution image inpainting from two aspects. On the one hand, most of the costly operations are in the low-resolution parameter generation network, where we can learn larger reception fields. On the other hand, we can decode the masked regions only since the coordinate can be queried one-by-one in parallel and the larger size image can be got via the coordinate re-sampling. The detailed experiments show the efficiency and the power of the proposed method  compared with previous state-of-the-art on several benchmarks.

The contributions of this paper are summarized as follows:

\begin{itemize}
    \item We propose CoordFill, a novel framework for efficient high-resolution image inpainting via parameterized coordinate querying.
    \item We design an attentional FFC-based block as the basic block in our parameter generation network. It learns to focus on the masked region automatically.
    \item Our method runs faster than previous baselines and achieves state-of-the-art performance in image inpainting on multiple datasets.
\end{itemize}

\section{Related Work}
\begin{figure*}[!t]
\centering
\includegraphics[width=\textwidth]{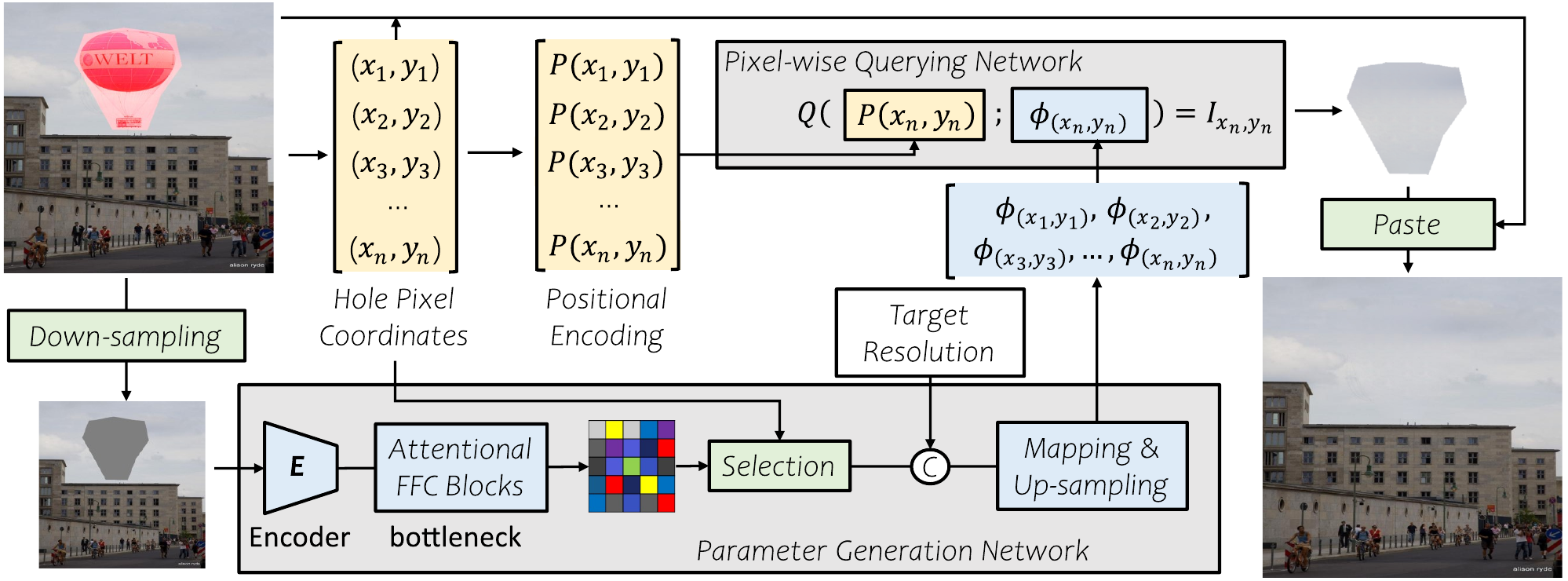}
\vspace{-2em}
\caption{Overview of the proposed framework.}
\label{overview}
\end{figure*}

\subsubsection{Image Inpainting}
Traditional image inpainting methods rely on strong low-level assumptions, for example, local patches~\cite{patchmatch} can be used to fill the missing region. Several recent CNN-based methods~\cite{liu2020rethinking,lama,global_and_local,deepfill,crfill,deepfillv2,Wan_2021_ICCV,pconv,rn, semantic_inpainting, structural_inpainting,mat,zits} have a similar or a stack of encoder-decoder architectures. More specifically, Iizuka~\emph{et.~al.}~\cite{global_and_local} introduce a GAN-based framework by the global and local discriminator. Based on the GAN framework, more novel blocks are also introduced through Attentions~\cite{deepfill,deepfillv2, liu2020rethinking,crfill}, Regional Normalization~\cite{rn} and Regional Convolutions~\cite{pconv}. Novel structures, like Vision Transformer~\cite{vit}, also draw the attention of the image editing community~\cite{Wan_2021_ICCV}. Meaningful priors, such as edge~\cite{edgeconnect} and semantic label~\cite{semantic_guide} also play an important role. For high-resolution image inpainting, a multi-stage network is a common choice. For example, Yi~\emph{et.~al.}~\cite{hifill} refine the low-resolution output via the contextual residual aggregation. Zeng~\emph{et.~al.}~\cite{zeng2020high} design a multi-scale network structure with guided up-sampling. MAT~\cite{mat} and ZITS~\cite{zits} are transformer-based image inpainting systems for high-resolution images. However, the multi-stage network structure is slow for real-world applications. In contrast, LaMa~\cite{lama} designs a one-stage network by the fusion of the multi-scale reception fields. However, their speed is also restricted by the normal decoder.

\subsubsection{Continuous Image Representation}
Learning the continuous image representation, \emph{i.e}, the implicit representation, are popular recently, due to the success in 3D human digitization~\cite{pifu, pifuhd} and novel view synthesis~\cite{nerf}. Generally, these methods query the color of the specific localization using the positional encoding in the frequency domain, which can generate more detailed high-frequency details than other methods. However, implicit representation is rarely developed in the 2D image domain. StyleGANv3~\cite{styleganv3} is built on implicit representation for alias-free settings of the image generation in the specific domain. InfinityGAN~\cite{infinitygan} also learns for image synthesis. ASAPNet~\cite{asapnet} learns the pixels from the semantic layout. LIIF~\cite{llif} learns a continuous representation for image super-solution. However, how to utilize this technique in image inpainting is still unclear and we make the first step.

\subsubsection{Efficient Network Structure}
Designing an efficient network structure for the high-resolution image is a more practical consideration for any deep learning-based method. Several attempts have also been made to make the convolutional network work in mobile applications. For example, the efficient network backbone has been widely explored in image classification~\cite{mobilenet,squeezenet}. In the image editing and synthesis fields, the models can be distilled from a trained larger network using knowledge distillation~\cite{gancompression}. For the image retouching tasks~\cite{3dlut, s2crnet}, learning the global-aware representation is helpful to keep the efficiency in high-resolution images. As for the high-resolution inpainting, efficiency is not the concern of the community yet.

\section{Method}
\label{sec:method}

\subsubsection{Overview}
\label{subsec:overview}

Our goal is to synthesize the missing content for high-resolution images in an efficient manner and hold a stable performance to diverse resolutions. To this end, our proposed model, dubbed as CoordFill, solves the inpainting problem with two phases: (i) restorative feature synthesis, as represented by spatially-adaptive parameters of the reconstruction function; (ii) individual pixel reconstruction based on the per-pixel coordinate query.
The advantage of this design is that we can synthesize high-resolution content efficiently for any specified regions thanks to the coordinate query-based reconstruction scheme. Therefore, on the one hand, it is allowed to selectively reconstruct the hole region of the input image. On the other hand, we can resample the input image with a fixed resolution before being fed to the network, which enables the contextual feature extraction under a unified receptive field, yet the image still could be restored with the original resolution (i.e. by sub-pixel coordinate query).

Specifically, given a high-resolution masked image $\mathbf{I}*\mathbf{M} \in \mathbb{R}^{H \times W \times 3}$ and its mask $\mathbf{M} \in [0, 1]^{H \times W}$, we first downsample them by $S$ times to $256 \times256$. 
Then, we build a Parameter Generation Network $G$ to generate the full image representation as a parameter map $\mathbf{\Phi}$. Finally, a Pixel-wise Querying Network $Q$, parameterized by $\mathbf{\Phi}_p$, generates pixel values according to the coordinate query $\mathbf{p}$ at the corresponding patches and finally generates the restored image $\mathbf{I}_o$ by pasting back the non-hole content.
The overall architecture is illustrated in Figure~\ref{overview}. Below, we give the details of each component.

\subsubsection{Parameter Generation Network}
\label{subsec:parameter_network}

Existing inpainting models are typically formulated to generate a full image with hole regions filled but only the hole region pixels are used in the final results. Obviously, the non-hole pixels are not necessary to be synthesized, as they contribute nothing to the final visualization. To avoid such unnecessary computation, we propose to first generate a parameter map from the input and then reconstruct the pixel values at hole regions individually.
As the inpainting task relies more on the surrounding context interpretation than texture details, we propose to resample the input with a fixed resolution before feeding it to our parameter generation network $G$. This operation not only promotes computation efficiency, especially for high-resolution images but also guarantees our network with good generalization to various input resolutions. Note that, we can still reconstruct the hole region with the original resolution, due to the coordinate query-based reconstruction scheme.

\begin{figure}[!t]
\centering
\includegraphics[width=\columnwidth]{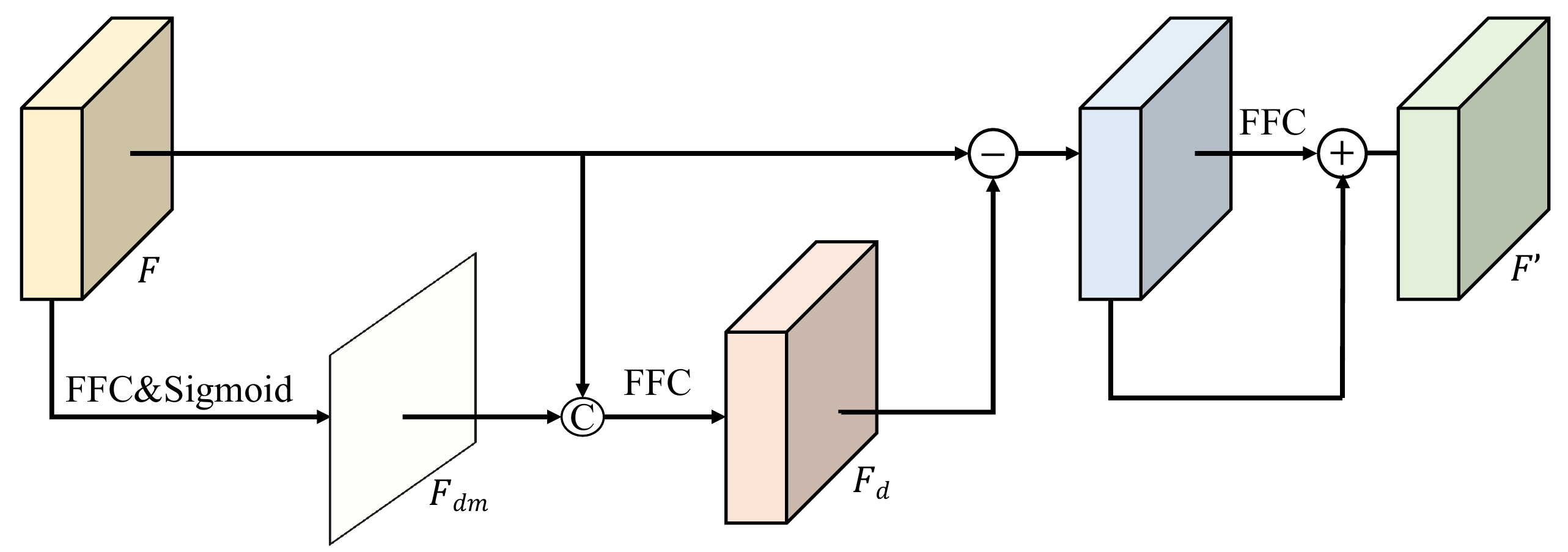}
\vspace{-2em}
\caption{The details of the proposed AttFFC Block.}
\label{efb}
\end{figure}

We employ a fully convolutional structure for the parameter generation network, composed of the encoder, bottleneck blocks, and the final layers of our parameter generation network.
In detail, we down-sample the input feature via the convolution layer three times to favor the process of calculation.  

In bottleneck blocks, inspired by recently popular Fast Fourier Convolution~(FFC)~\cite{ffc,lama} blocks, we propose an FFC-based framework also. This block captures the global reception fields of the input in the frequency domain, yet a better performance in image inpainting which needs larger reception fields. Please refer to the previous works~\cite{ffc,lama} for more details.

However, the global receptive field in FFC is only limited in the frequency domain, which lacks representation in the spatial space.
To alleviate this problem, we propose a spatial-aware attention block based on FFC~(AttFFC). 
As shown in Figure~\ref{efb}, the proposed block contains $-/+$ operations to remove the spatial noise and enhance the features~\cite{distraction-aware}.
Given a feature map $\mathbf{F} \in \mathbb{R}^{H \times W \times C}$, AttFFC first computes the spatial attention map $\mathbf{F}_{dm}$ through a FFC Block and a Sigmoid function $\sigma$:
\begin{equation}
\mathbf{F}_{dm} = \sigma(\operatorname{FFC}(\mathbf{F})).
\end{equation}
where $F_{dm}$ are the one channel spatial confidential map automatically. Next, we use this spatial prior and the original feature to get the spatial noise:
\begin{equation}
\mathbf{F}_{d} = \operatorname{FFC}\left(\operatorname{concat}\left(\mathbf{F}, \mathbf{F}_{dm}\right)\right).
\end{equation}
Then, we remove the spatial noise, and enhance the features by learning the residual with an FFC block:
\begin{equation}
\mathbf{F}'= \mathbf{F} - F_{d} + \operatorname{FFC}(\mathbf{F} - \mathbf{F}_{d}).
\end{equation}

We stack six AttFFCs in our parameter generation network as the bottleneck.

Notice that, since we use the encoder to reduce the feature space, the learned spatial feature is also in a small resolution. To this end, we build the pixel-wise parameters via selection, mapping, and upsampling. In detail, we first select the masked region features by the corresponding coordinates, and then, for each spatial feature, we generate the required number of the parameters by the linear mapping function $f$. In addition, to make the parameter generation network sensitive to the target resolution $r$, we inject the target resolution as a conditional input to $f$:
\begin{equation}
\phi_{m} = f(cat[F_{m}, r]),
\end{equation}
where $F_m$ is the features in the masked region only, and $cat[\cdots]$ denotes the concatenation operation.
Finally, the spatial-adaptive patch $\phi_{m}$ will be upsampled to the required resolution with nearest-neighbor interpolation. Hence, we obtain a series of parameter vectors with channels corresponding to the weights~(and the biases) of the MLPs, and the heavy computation is performed at low resolution.

\subsubsection{Pixel-wise Querying Network}
\label{subsec:query_network}

As mentioned above, we adopt a coordinate query-based pixel reconstruction scheme by utilizing the spatially-adaptive parameters. In particular, inspired by the efficiency of implicit representation~\cite{asapnet}, we employ a Multi-Layer Perception~(MLP) for per-pixel reconstruction, \emph{i.e.} the Pixel-wise Querying Network $Q$, which is parameterized by the spatial-adaptive contextual information embedded parameters, as extracted by $G$. Note that, $Q$ is just a function form and its expressive capability mainly inherits from those deeply extracted parameters $\mathbf{\Phi}$. Formally, $Q$ takes the pixel's coordinate $\mathbf{p}$ as input and outputs the color values:
\begin{equation}
\mathbf{y}_{p} = Q(\mathbf{p};\mathbf{\Phi}_{p}),
\end{equation}
To boost high-frequency details, we encode each component of the 2D pixel position $p = (P_x, P_y)$ as a vector of sinusoids~\cite{transformer}.
Additionally, we normalize the positional encoding system and sample positional encoding within a fixed range. Thus, the pixels in the high-resolution images can be synthesized by changing the interval of the positional encoding. For example, given an input image $\mathbf{I}$ and its hole mask $\mathbf{M}$ of resolution $H \times W$, our parameter generation network generates the spatially-adaptive parameter map $\mathbf{\Phi}$ of resolution $h \times w$ from the downsampled input $\{\tilde{\mathbf{I}}, \tilde{\mathbf{M}}\}$. Then, we feed the coordinate query:
\begin{equation}
\begin{aligned}
\mathbf{p} = (\sin \left(2 \pi p_{x} / \mathbf{E}_{x}\right), \cos \left(2 \pi p_{x} / \mathbf{E}_{x}\right), \\
\sin \left(2 \pi p_{y} / \mathbf{E}_{y}\right), \cos \left(2 \pi p_{y} / \mathbf{E}_{y}\right))
\end{aligned}
\end{equation}
at expected interval $\mathbf{E}_{x}$, $\mathbf{E}_{y}$, where $\mathbf{E}_{x} = \frac{H}{h}$ and $\mathbf{E}_{y} = \frac{W}{w}$
, and then synthesize the hole-region content of the original resolution.

\subsubsection{Loss Function}
\label{sec:loss}
Since image inpainting only needs to measure the realism of the content, we train our model with different perception losses, including the differences in the pre-trained AlexNet domain~\cite{lpips,jo2020investigating}, adversarial loss~\cite{gan}, and feature matching loss~\cite{lama,pix2pixhd}.

For the perceptual loss on the pre-trained ImageNet classification domain, the loss function is defined as:
\begin{equation}
L_{per}=\sum_{k} \tau_{k}\left(E_{k}\left(\mathbf{I}_o\right)-E_{k}\left(\mathbf{I}_{gt}\right)\right),
\end{equation}
where $E$ is an AlexNet feature extractor, $\tau$ calculates the differences in the feature domain via $L_1$ loss, and then, we compute and average the losses from $k$ layers.

Then, the adversarial loss is used to encourage the model to generate more realistic details, which can be written as:
\begin{equation}
\begin{aligned}
L_{adv} = \mathbb{E}[\log (1-D(\mathbf{I}_o)] + \mathbb{E}[\log D(\mathbf{I}_{gt})].
\end{aligned}
\end{equation}
where $\mathbb{E}$ denotes expectation values over the training batch.

Next, the feature matching loss is adopted for stabilizing the GAN training~\cite{pix2pixhd,lama}:
\begin{equation}
L_{fm} = \sum_{i}\left(D^{i}\left(\mathbf{I}_{gt}\right), D^{i}\left(\mathbf{I}_o\right)\right),
\end{equation}
where $D^{i}$ denotes the activations from the $i$-th layer of the discriminator D.

Finally, the total loss of our model can be written as:
\begin{equation}
L_{total} = \lambda_{per} L_{per} + \lambda_{adv} L_{adv} + \lambda_{fm} L_{fm}.
\end{equation}
where we empirically set $\lambda_{per}=10$, $\lambda_{adv}=1$ and $\lambda_{fm}=100$ respectively.

\begin{figure*}[!t]
\centering
\includegraphics[width=\textwidth]{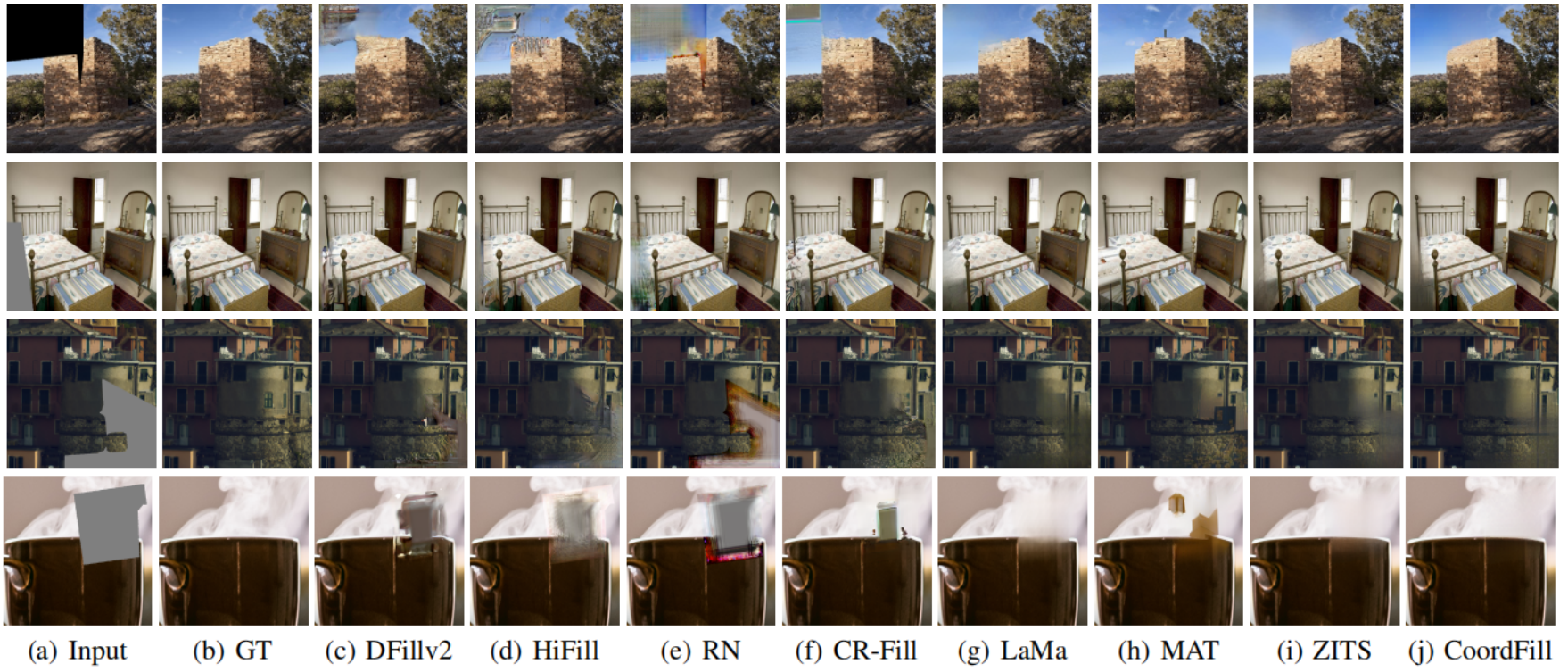}
\vspace{-2em}
\caption{Qualitative comparisons of the proposed CoordFill and other methods. The top two examples are 512$\times$512 images from the Places2 dataset while the bottom two images are 1024$\times$1024 images from the Unsplash dataset.}
\label{fig:sota_places2}
\vspace{-1em}
\end{figure*}

\begin{table*}[!t]
\caption{Comparison with state-of-the-art methods on Places2 Dataset on different resolutions. ``-" means these methods cause the out-of-memory issue in the given experiment resolutions. }
\vspace{-1em}
\centering  
\resizebox{\textwidth}{!}{
\begin{tabular}{l|llll|llll|llll|llll}
\toprule
 Resolution
 & \multicolumn{4}{c|}{512$\times$512} &
  \multicolumn{4}{c|}{1024$\times$1024} &
  \multicolumn{4}{c|}{2048$\times$2048} &
  \multicolumn{4}{c}{4096$\times$4096} \\ \hline
 &
 \multicolumn{1}{c}{PSNR$\uparrow$} &
  \multicolumn{1}{c}{SSIM$\uparrow$} &
  \multicolumn{1}{c}{LPIPS$\downarrow$} &
  \multicolumn{1}{c|}{SPEED$\downarrow$} &
  \multicolumn{1}{c}{PSNR$\uparrow$} &
  \multicolumn{1}{c}{SSIM$\uparrow$} &
  \multicolumn{1}{c}{LPIPS$\downarrow$} &
  \multicolumn{1}{c|}{SPEED$\downarrow$} &
  \multicolumn{1}{c}{PSNR$\uparrow$} &
  \multicolumn{1}{c}{SSIM$\uparrow$} &
  \multicolumn{1}{c}{LPIPS$\downarrow$} &
  \multicolumn{1}{c|}{SPEED$\downarrow$} &
  \multicolumn{1}{c}{PSNR$\uparrow$} &
  \multicolumn{1}{c}{SSIM$\uparrow$} &
  \multicolumn{1}{c}{LPIPS$\downarrow$} &
  \multicolumn{1}{c}{SPEED$\downarrow$} \\ \hline
DeepFillv2~\cite{deepfillv2} &
23.973 & 0.902 & 0.080 & 398ms &
22.695 & 0.908 & 0.092 & 1002ms &
-& -& -& -&
-& -& -& -\\
HiFill~\cite{hifill} &
23.375 & 0.883 & 0.097 & 406ms &
23.456 & 0.894 & 0.096 & 423ms &
23.643 & 0.915 & 0.087 & 478ms &
23.634 & 0.933 & 0.077 & 662ms \\
RN~\cite{rn} &
22.562 & 0.880 & 0.116 & 17ms &
19.587 & 0.879 & 0.139 & 59ms &
18.843 & 0.908 & 0.143 & 240ms &
-& -& -& -\\
CR-Fill~\cite{crfill} &
24.216 & 0.893 & 0.086 & 46ms &
22.881 & 0.890 & 0.108 & 54ms &
22.056 & 0.908 & 0.122 & 63ms &
-& -& -& -\\
LaMa~\cite{lama} &
26.203 & \textbf{0.914} & \textbf{0.067} & 27ms &
26.154 & \textbf{0.924} & 0.076 & 142ms &
25.688 & \textbf{0.939} & 0.078 & 598ms &
-& -& -& -\\
MAT~\cite{mat} &
24.169 & 0.900 & 0.076 & 71ms &
23.751 & 0.908 & 0.082 & 133ms &
- & - & - & -&
-& -& -& -\\
ZITS~(Dong et al. 2022) &
26.349 & 0.911 & 0.068 & 183ms &
\textbf{26.389} & 0.913 & \textbf{0.073} & 462ms & 
- & - & - & - &
-& -& -& -\\
\textbf{CoordFill} &
\textbf{26.365} & \underline{0.912} & \underline{0.068} & \textbf{10ms} &
\underline{26.322} & \underline{0.920} & \underline{0.075} & \textbf{14ms} &
\textbf{26.322} & \underline{0.932} & \textbf{0.077} & \textbf{26ms} &
\textbf{26.175} & \textbf{0.943} & \textbf{0.075}  & \textbf{78ms}\\ \hline
\end{tabular}}
\label{tab:compare}
\vspace{-1em}
\end{table*}

\begin{table}[t]
\caption{Comparison with state-of-the-art methods on Unsplash and CelebA-HQ Datasets~(resolution 1024$\times$1024). 
}
\vspace{-1em}
\resizebox{\columnwidth}{!}{
\centering
\begin{tabular}{l|l|l|l|l|l}
\toprule
    & Methods
    & PSNR$\uparrow$ & SSIM$\uparrow$ & LPIPS$\downarrow$  & 
    SPEED$\downarrow$ \\ \hline
\parbox[t]{2mm}{\multirow{8}{*}{\rotatebox[origin=c]{90}{Unsplash}}}
\multirow{4}{*}
&  DeepFillv2~\cite{deepfillv2}
&  24.859 &  0.914 &  0.097 
& 1002ms
\\
&  HiFill~\cite{hifill}
&  27.653 &  0.914 &  0.082
&  423ms
\\
&  RN~\cite{rn}
&  20.290 &  0.890 &  0.162  
& 59ms
\\
& CR-Fill~\cite{crfill}
&  28.690  &  0.923 &  0.082
& 54ms
\\
& LaMa~\cite{lama}
&  32.434 & 0.934 & 0.055
& 142ms
\\
& MAT~\cite{mat}
&  29.506 &  0.923 &  0.081
& 133ms
\\
& ZITS~\cite{zits}
&  32.532 &  0.934 &  \textbf{0.053}
& 462ms
\\
& \textbf{CoordFill}
&  \textbf{33.093} &  \textbf{0.934} &  0.080  &  \textbf{14ms}
\\ \hline
\parbox[t]{2mm}{\multirow{4}{*}{\rotatebox[origin=c]{90}{
\footnotesize{CelebA-HQ}
}}}
&  DeepFillv2~\cite{deepfillv2}
&  23.286 &  0.916 &  0.106  &  1002ms
\\
&  LaMa~\cite{lama}
&  26.098 &  0.924 &  0.080
& 142ms
\\
& MAT~\cite{mat}
&  25.167 & 0.917  &  0.074
&  133ms
\\
& \textbf{CoordFill}
&  \textbf{28.756} &  \textbf{0.934} &  \textbf{0.065} & \textbf{14ms}
\\
\bottomrule
\end{tabular}}
\label{tab:sota_highres}
\vspace{-1em}
\end{table}

\section{Experiments}
\label{sec:exp}

\subsection{Implementation Details}
\label{sec:Implementation Details}
We evaluate our method on the widely-used Places2~\cite{place_dataset} dataset, as well as two high-resolution image datasets, CelebA-HQ~\cite{progan} and Unsplash~\cite{unsplash}. We use the irregular mask from the previous method~\cite{pconv} for mask generation with holes up to 25\% following HiFill~\cite{hifill}.  
Our model is optimized by the Adam optimizer with a learning rate of 1e-4 and trained with 100 epochs. All the training experiments are conducted on 8 Tesla A100 GPUs with a batch size of 128. All the images are randomly resized from 256$\times$256 to 512$\times$512 during training, then, we test them in the range of 512, 1024, 2048, and 4096. Following previous methods~\cite{hifill,aotgan}, we use the PSNR, SSIM, and LPIPS~\cite{lpips} for performance comparison.

\subsection{Comparisons with Prior Arts}
\label{sec:State-of-the-art Comparisons}
Our method is compared against the existing state-of-the-art image inpainting algorithms including the general image inpainting models~(DeepFillv2~\cite{deepfillv2},  RN~\cite{rn} and CR-Fill~\cite{crfill}), and the high-resolution image inpainting models~(HiFill~\cite{hifill}, LaMa~\cite{lama}, MAT~\cite{mat}, and ZITS~\cite{zits}) using their official pre-trained models.
As shown in Table~\ref{tab:compare}, our CoordFill achieves competitive performance at all resolutions on the Places2 validation set. Compared with the common image inpainting methods~(DeepFillv2, RN, and CR-Fill) on the 512$\times$512 images, the proposed method achieves the best performance. Compared with the existing high-resolution image inpainting methods~(HiFill, LaMa, MAT, and ZITS) on various resolutions, the proposed method also shows more stable results when the resolution increases thanks to the robust resolution and continuous pixel-wise query. For visual comparison, we also compare our methods with other state-of-the-art methods in Figure~\ref{fig:sota_places2}. According to the figure, the proposed method shows more visual-please results than the previous methods with stable reception fields and pixel-wise querying.

Since the previous high-resolution image inpainting method~\cite{hifill} generates the high-resolution samples~($>$512) by the bilinear upsampling on Places2, we use another two high-resolution image datasets to evaluate the performance of each method more accurately, including CelebA-HQ and Unsplash. CelebA-HQ is a commonly used high-resolution face dataset with a resolution of 1024$\times$1024, and the model is retrained with its train set and evaluated on its test set. Unsplash is a real-world high-resolution dataset collected from Unsplash~\cite{unsplash} in which the original images are cropped to 1024$\times$1024, and we use the pre-trained model on Place2 for evaluation. Table~\ref{tab:sota_highres} reports the performance of each method on these datasets, where CoordFill shows competitive performance with the fastest inference speed. As for the visual comparison, the proposed CoordFill also generates clear details as in Figure~\ref{fig:sota_places2}.

\subsection{Efficiency Analysis}
We give a detailed efficiency analysis. Firstly, we show the speed comparison of our method and others in Table~\ref{tab:compare}. The speed is calculated using the average inference time on an NVIDIA GTX 2080 Ti GPU. Specifically, the proposed method runs much faster than all the methods under all the resolutions.
On the same experiment platform, nearly all the methods cause out-of-memory memory issues when the resolution increases to 4096$\times$4096 except ours and HiFill. Also, as shown in Table~\ref{tab:compare}, the proposed method shows a much better performance. 
\begin{figure}[t]
\centering
\includegraphics[width=\columnwidth]{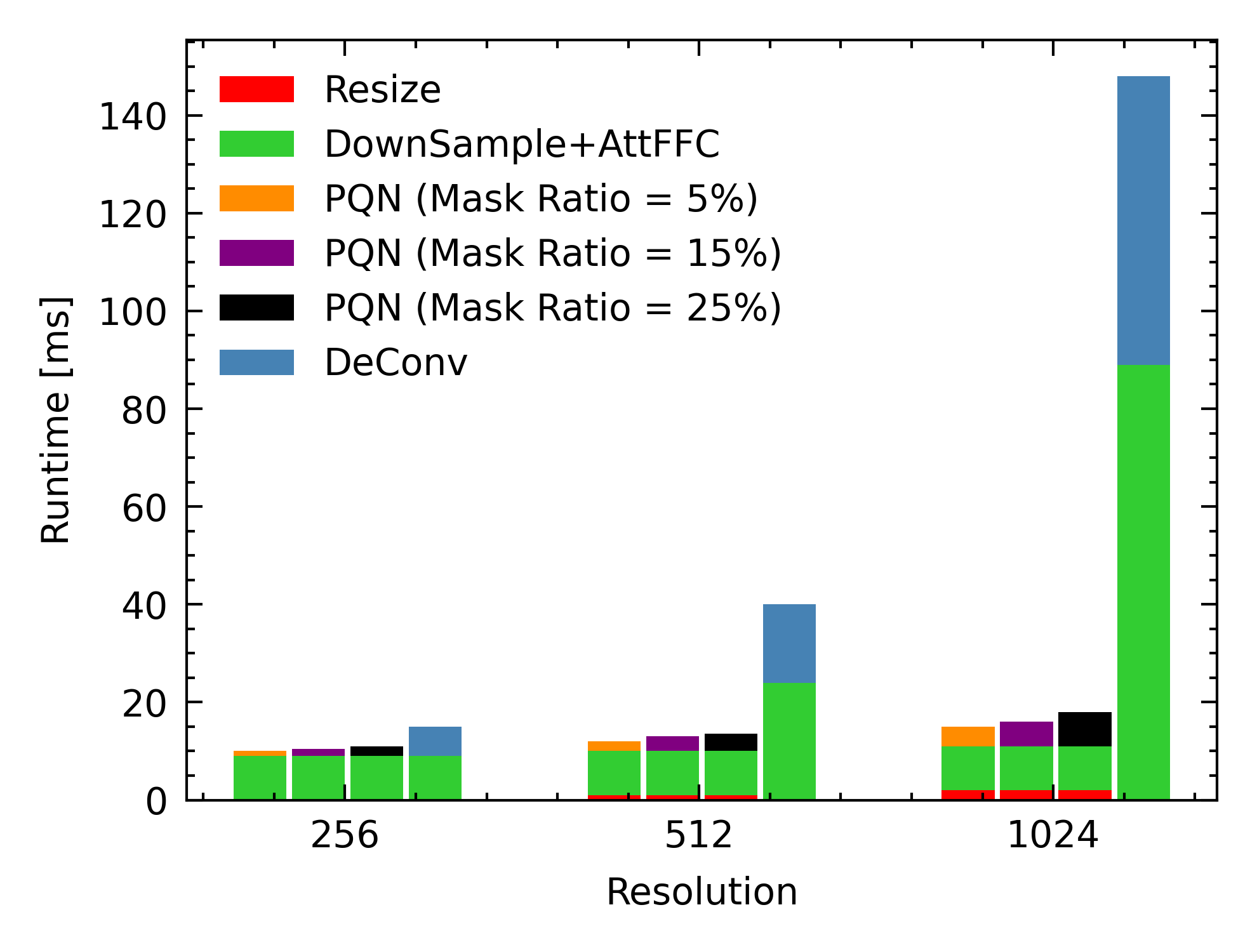}
\vspace{-2.5em}
\caption{The speed comparison of the proposed method~(three different mask ratios) and the baseline~(DownSample + AttFFC + $D_{Conv}$) on the three different resolutions.}
\vspace{-1em}
\label{component_speed}
\end{figure}

Furthermore, as shown in Figure~\ref{component_speed}, we give a detailed explanation of what exactly influences the speed when the resolution changes. As in Figure~\ref{component_speed}, our pixel-wise querying network always runs faster than the normal convolutional decoder~($D_{Conv}$), On the other hand, the speedup comes from the stable reception fields where larger feature maps will speed more time. It is also interesting to see that the running time of our method is flexible when the hole changes. It is because our method only needs to predict the masked region using the pixel-query strategy.

\begin{table}[t]
\caption{Ablation study on the Places2 dataset.}
\vspace{-1em}
\resizebox{\columnwidth}{!}{
\centering
\begin{tabular}{c|c|c|c|ccc}
\toprule
    Masked
    & \multirow{2}{*}{Decoder} 
    & \multirow{2}{*}{Block}  
    & Resolution
    &  \multirow{2}{*}{PSNR$\uparrow$} 
    & \multirow{2}{*}{SSIM$\uparrow$}   
    & \multirow{2}{*}{LPIPS$\downarrow$}   \\ 
    Prediction
    & 
    & 
    & Injection
    & 
    \\ \hline
    
-
&  $D_{Conv}$
&  ResFFC & -
&  25.5420 &  0.9074 &  0.0688
\\
$\checkmark$
&  $D_{Conv}$
&  ResFFC & -
&  25.9557 &  0.9097 &  0.0679
\\
$\checkmark$
&  $D_{MLP}$
&  ResFFC & -
&  24.6480 &  0.8995 &  0.1186  
\\
-
&  Pixel-Query
&  ResFFC & -
&  26.0434  &  0.9091 &  0.0683
\\
$\checkmark$
&  Pixel-Query
&  ResFFC & -
&  26.2160 & 0.9109 & 0.0679
\\
$\checkmark$
&  Pixel-Query
&  AttFFC
& -
&  26.3177 &  0.9116 &  0.0679
\\
\textbf{$\checkmark$}
&  \textbf{Pixel-Query}
&  \textbf{AttFFC}
& \textbf{$\checkmark$}
&  \textbf{26.3653} &  \textbf{0.9119} &  \textbf{0.0677}
\\ \hline

\end{tabular}}
\label{tab:ablation}
\vspace{-1em}
\end{table}
%

\subsection{Ablation Study}
\label{sec:Ablation Study}
We conduct a detailed ablation study to verify the efficiency of each component in the proposed framework on the 512$\times$512 Places2 images, and the quantitative results are shown in Table~\ref{tab:ablation}.

\paragraph{\textbf{Supervision strategy}}
We design a novel framework for image inpainting, \emph{i.e.} selective region synthesis based on the coordinate query. Interestingly, we find that existing methods still calculate the loss function on the whole image, which might be biased supervision because we only care about the hole region accuracy. To verify this speculation, we calculate the loss function in the masked region only, called \textit{Masked Prediction}. As shown in Table~\ref{tab:ablation}, calculating the loss function in the masked region improves the performance in both $D_{Conv}$ and the proposed pixel-wise querying network structure on both two datasets.  It might be because the original supervision strategy for the full image synthesis will be influenced by the extra reception fields and result in color bleeding artifacts as shown in Figure~\ref{fig:mask_predict}.

\paragraph{\textbf{Importance of the coordinate-query scheme}}
The proposed pixel-wise querying network helps our method to obtain high-resolution and high-quality results, which is the key to our method. Thus, we compare it with $D_{Conv}$ and $D_{MLP}$ with the same encoder, where $D_{Conv}$ uses the original convolutional decoder for up-sampling and $D_{MLP}$ feed the pixel-wise feature to a shared MLP for pixel value decoding.
As shown in Table~\ref{tab:ablation} and Figure~\ref{fig:ablation}, the proposed pixel-wise querying framework gets much better performance and runs faster than the previous.

\newcommand\wwmp{0.3\columnwidth}
\newcommand\hhmp{0.3\columnwidth}

\begin{figure}[t]
\centering

\subfigure{\centering\includegraphics[width=\wwmp, height=\hhmp]{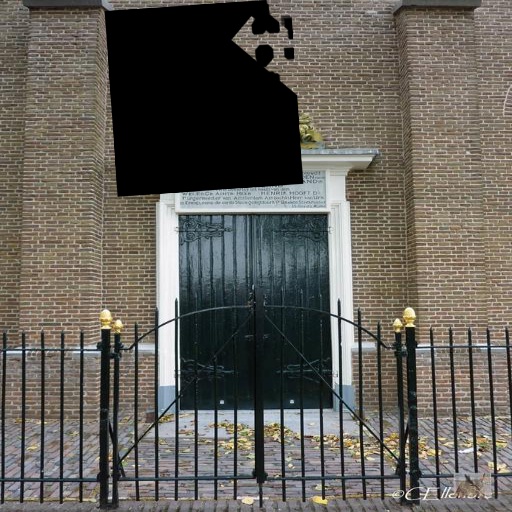}}
\subfigure{\centering\includegraphics[width=\wwmp, height=\hhmp]{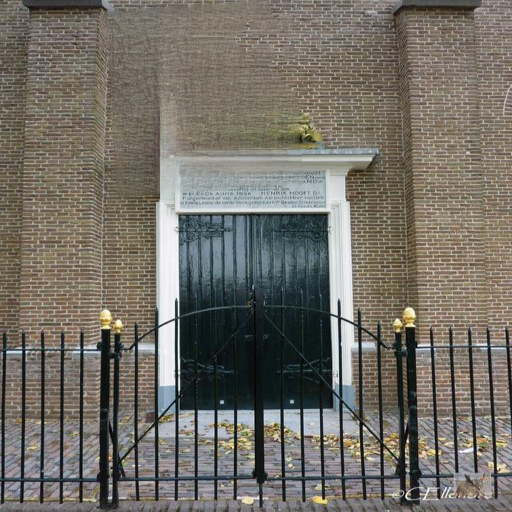}}
\subfigure{\centering\includegraphics[width=\wwmp, height=\hhmp]{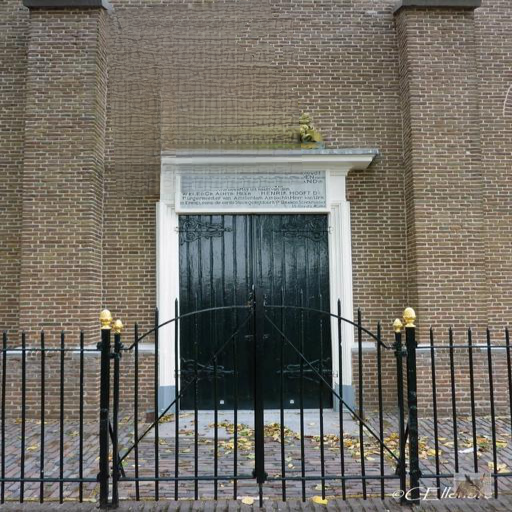}}

\par\bigskip\vspace*{-2em}
\subfigure{\centering\includegraphics[width=\wwmp, height=\hhmp]{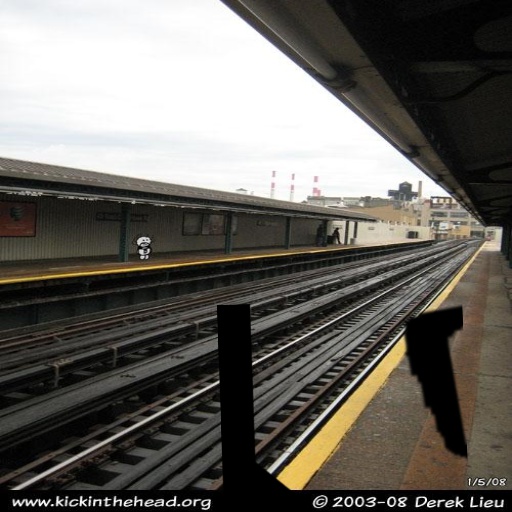}}
\subfigure{\centering\includegraphics[width=\wwmp, height=\hhmp]{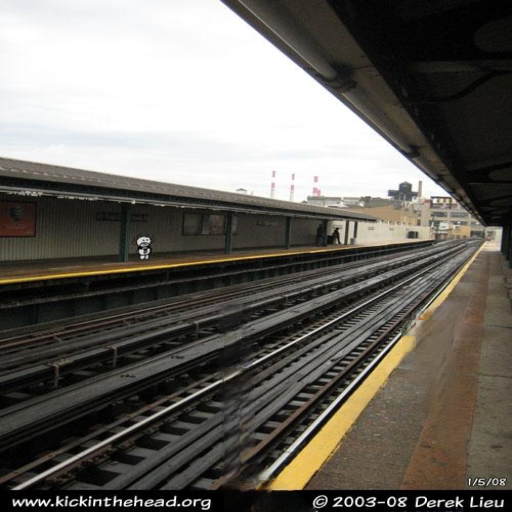}}
\subfigure{\centering\includegraphics[width=\wwmp, height=\hhmp]{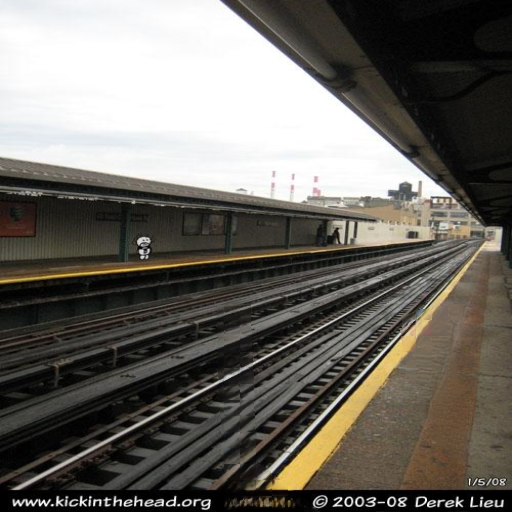}}

\par\bigskip\vspace*{-2em}
\addtocounter{subfigure}{-6}
\subfigure[\scriptsize{Input}]{\centering\includegraphics[width=\wwmp, height=\hhmp]{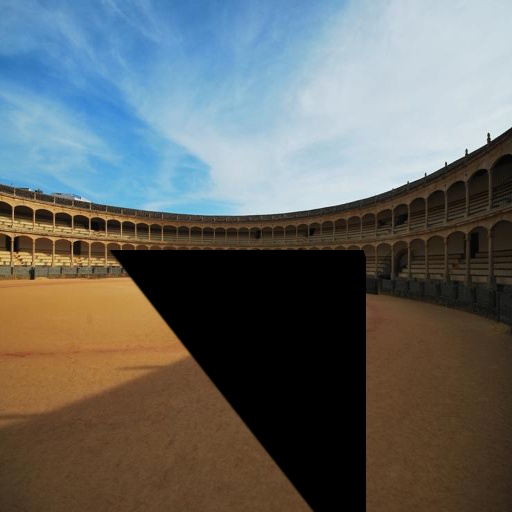}}
\subfigure[\scriptsize{w/o masked pred.}]{\centering\includegraphics[width=\wwmp, height=\hhmp]{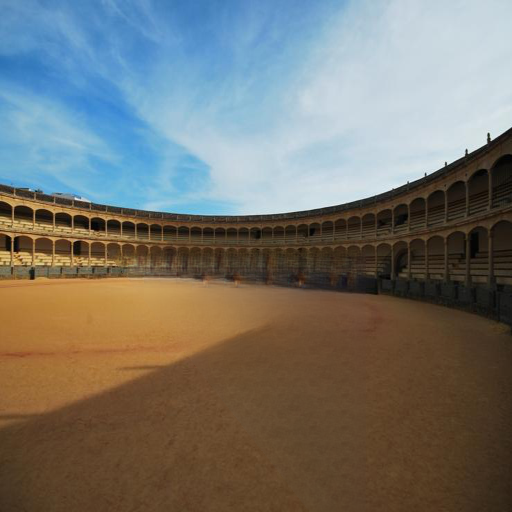}}
\subfigure[\scriptsize{w/ masked pred.}]{\centering\includegraphics[width=\wwmp, height=\hhmp]{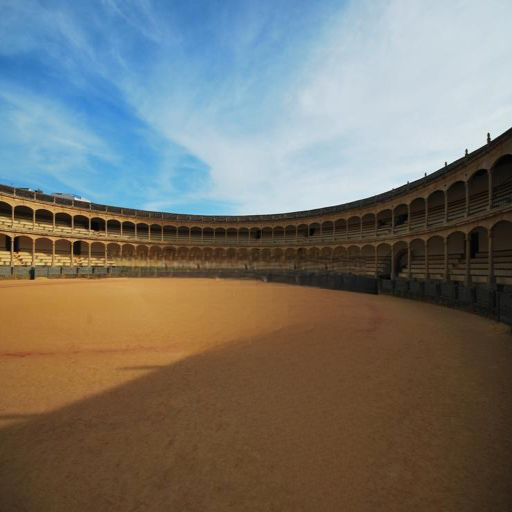}}

\vspace{-1em}
\caption{Qualitative comparisons of different supervision strategies.}
\label{fig:mask_predict}
\end{figure}

\newcommand\wwab{0.24\columnwidth}
\newcommand\hhab{0.24\columnwidth}
\begin{figure}[h!]
\centering

\subfigure{\centering\includegraphics[width=\wwab, height=\hhab]{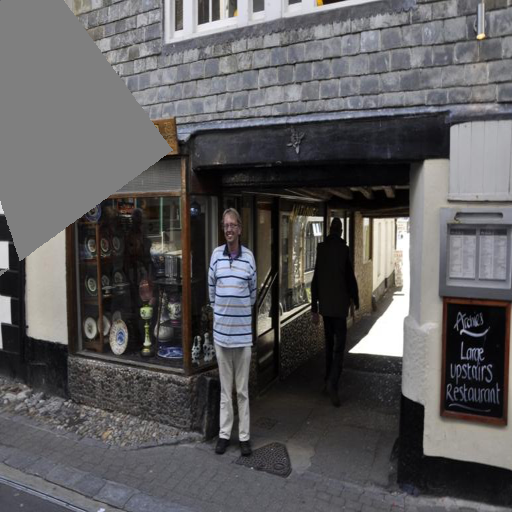}}
\subfigure{\centering\includegraphics[width=\wwab, height=\hhab]{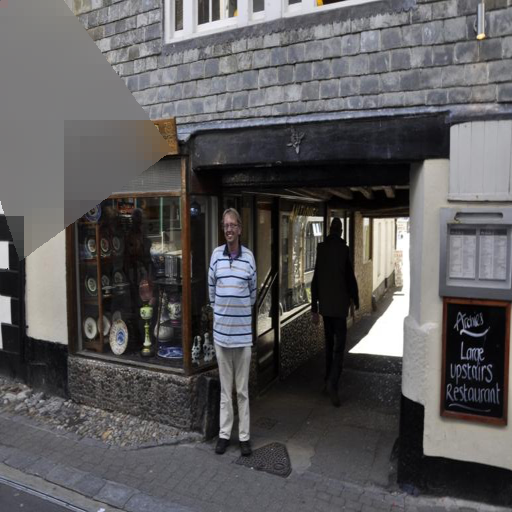}}
\subfigure{\centering\includegraphics[width=\wwab, height=\hhab]{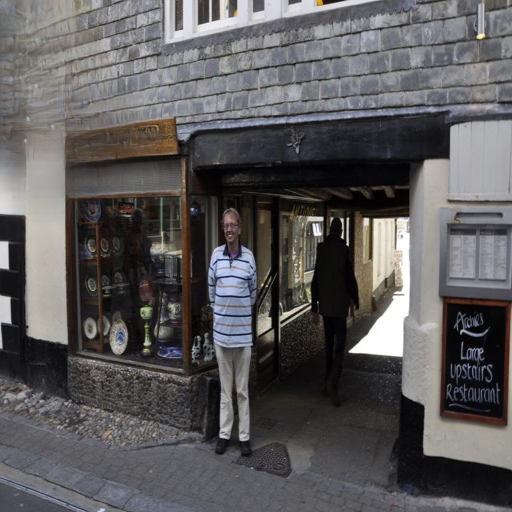}}
\subfigure{\centering\includegraphics[width=\wwab, height=\hhab]{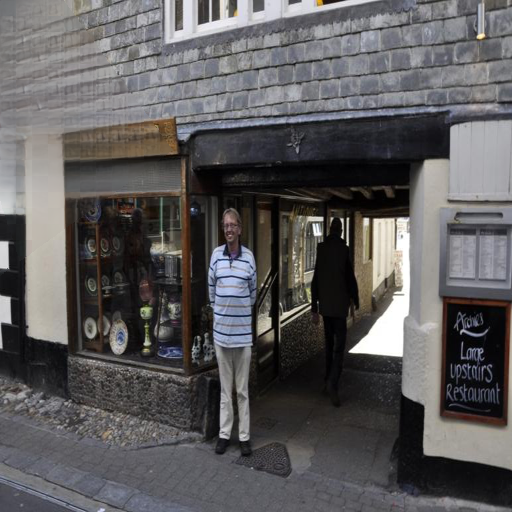}}

\par\bigskip\vspace*{-2em}
\subfigure{\centering\includegraphics[width=\wwab, height=\hhab]{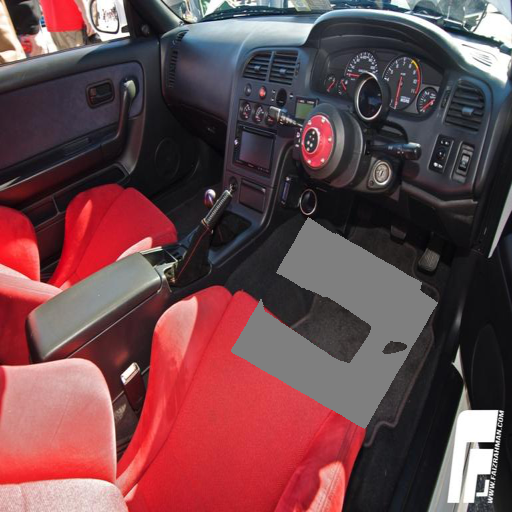}}
\subfigure{\centering\includegraphics[width=\wwab, height=\hhab]{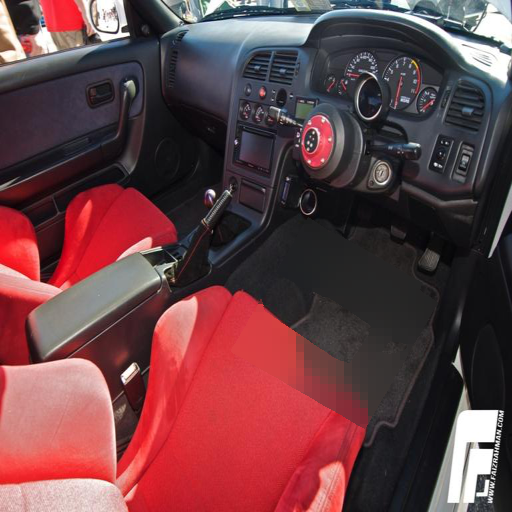}}
\subfigure{\centering\includegraphics[width=\wwab, height=\hhab]{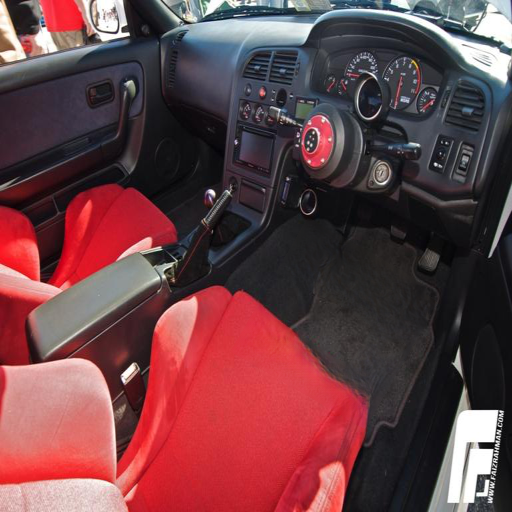}}
\subfigure{\centering\includegraphics[width=\wwab, height=\hhab]{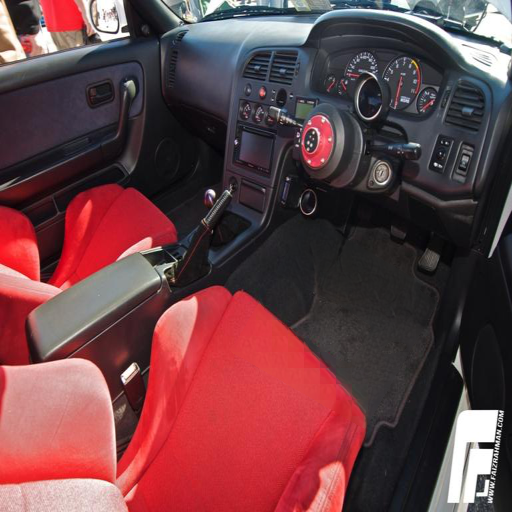}}

\par\bigskip\vspace*{-2em}
\addtocounter{subfigure}{-8}
\subfigure[\scriptsize{Input}]{\centering\includegraphics[width=\wwab, height=\hhab]{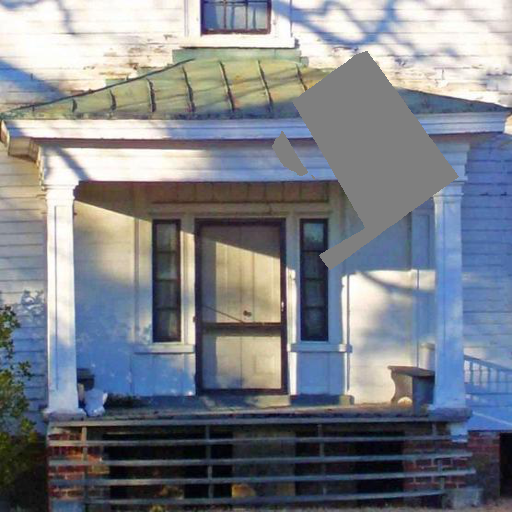}}
\subfigure[\scriptsize{w/ $D_{MLP}$}]{\centering\includegraphics[width=\wwab, height=\hhab]{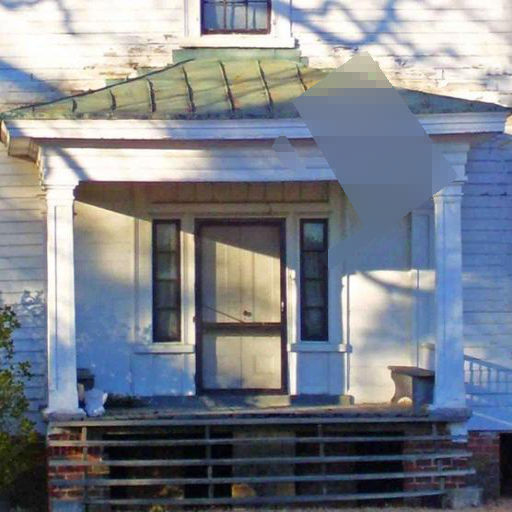}}
\subfigure[\scriptsize{w/ $D_{Conv}$}]{\centering\includegraphics[width=\wwab, height=\hhab]{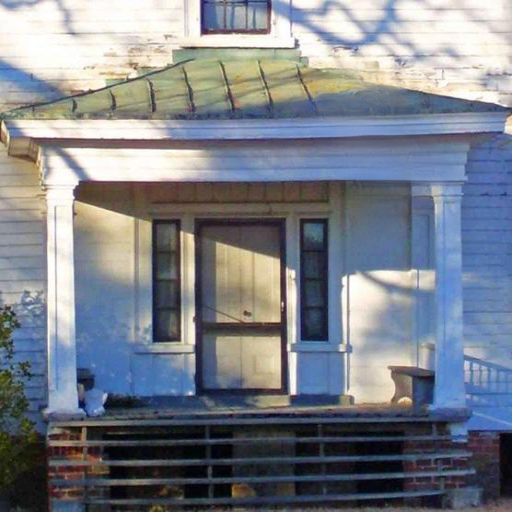}}
\subfigure[\scriptsize{w/ Pixel-Query}]{\centering\includegraphics[width=\wwab, height=\hhab]{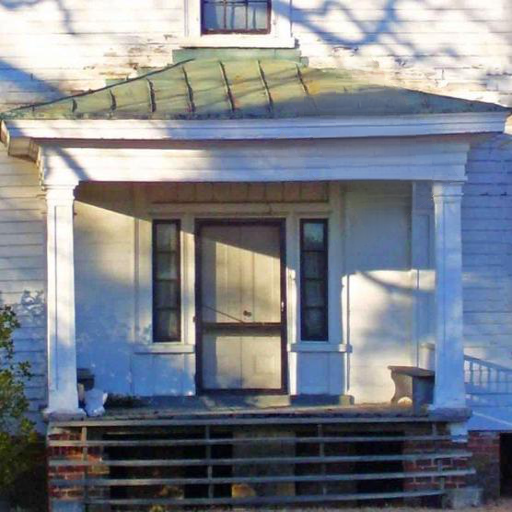}}

\vspace{-1em}
\caption{Qualitative comparisons of decoding using a shared MLP, CNN, and the pixel-wise querying network.}
\vspace{-1em}
\label{fig:ablation}
\end{figure}


\paragraph{\textbf{Advantage of our attentional FFC}}

The proposed attentional FFC~(AttFFC) is also an important component. As we have claimed, in AttFFC, we suppress the spatial noise by filtering out the distracting features. Compared with the original FFC blocks~(ResFFC) in LaMa~\cite{lama}, AttFFC achieves a better result than ResFFC as shown in Table~\ref{tab:ablation}.

\paragraph{\textbf{Impact of resolution injection}}
To extract features under a unified receptive field, the parameter generation network accepts a fixed low-resolution input. Hence the pixel-wise querying network generates the same results for different resolutions, and the high-frequency information can only be synthesized with the help of the input positional encoding. By condition on the resolution information, our method got a better performance as shown in Table~\ref{tab:ablation}.

\paragraph{\textbf{Comparisons with inpainting then super-resolution}}
Our coordinate query-based decoding network supports synthesizing the content in arbitrary resolution, and it is also implicitly required to consider the super-resolution as post-processing during testing. To check the effectiveness, we construct a baseline: the decoding network only generates the output of the same resolution as the down-sampled input~(256$\times$256) and then gets high-resolution image~(1024$\times$1024 in our case) by an extra up-sampling or super-resolution network.
We plot the visual results to support our claim in Figure~\ref{fig:super-resolution}, where the proposed method shows sharper and visual-friendly results.

\newcommand\wwsr{0.15\columnwidth}
\newcommand\hhsr{0.15\columnwidth}
\begin{figure}[t]
\captionsetup[subfigure]{labelformat=empty}
\centering  
\subfigure{\centering\includegraphics[width=\wwsr, height=\hhsr]{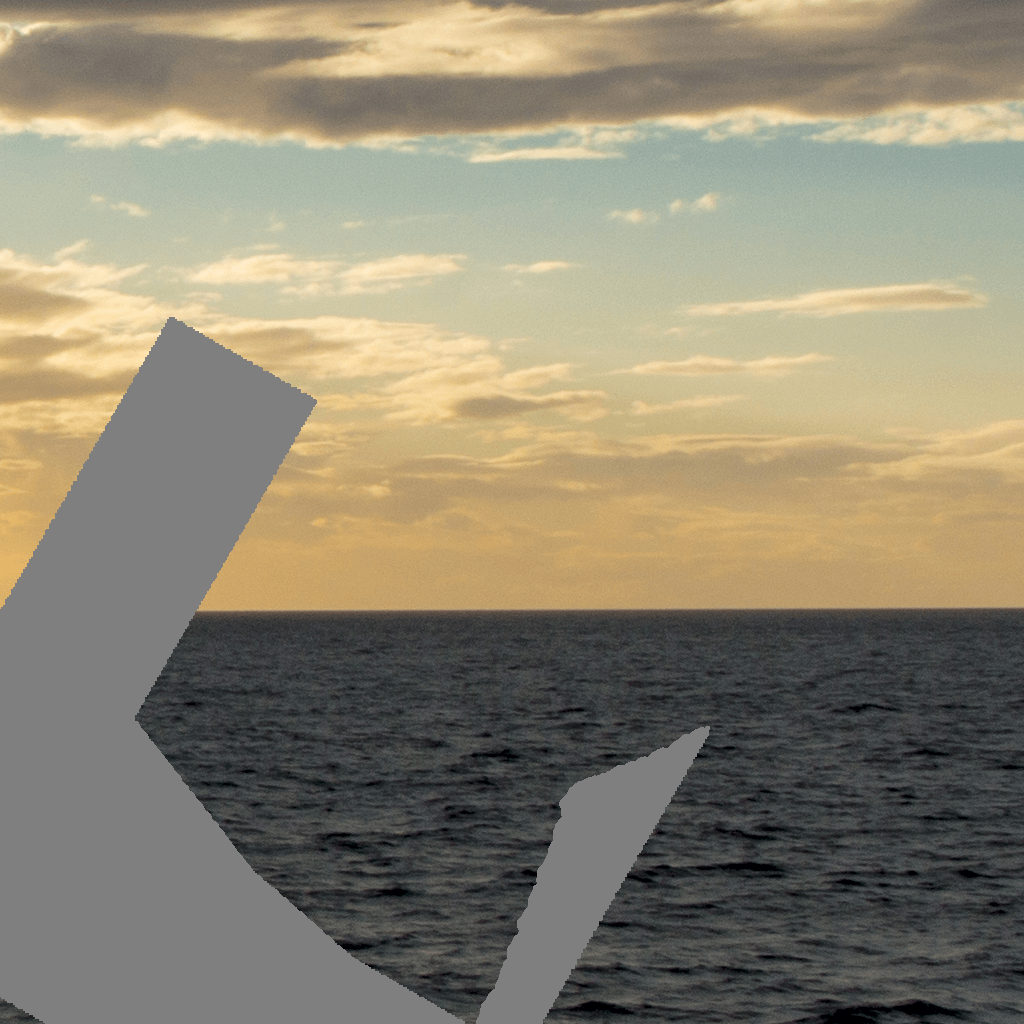}}
\subfigure{\centering\includegraphics[width=\wwsr, height=\hhsr]{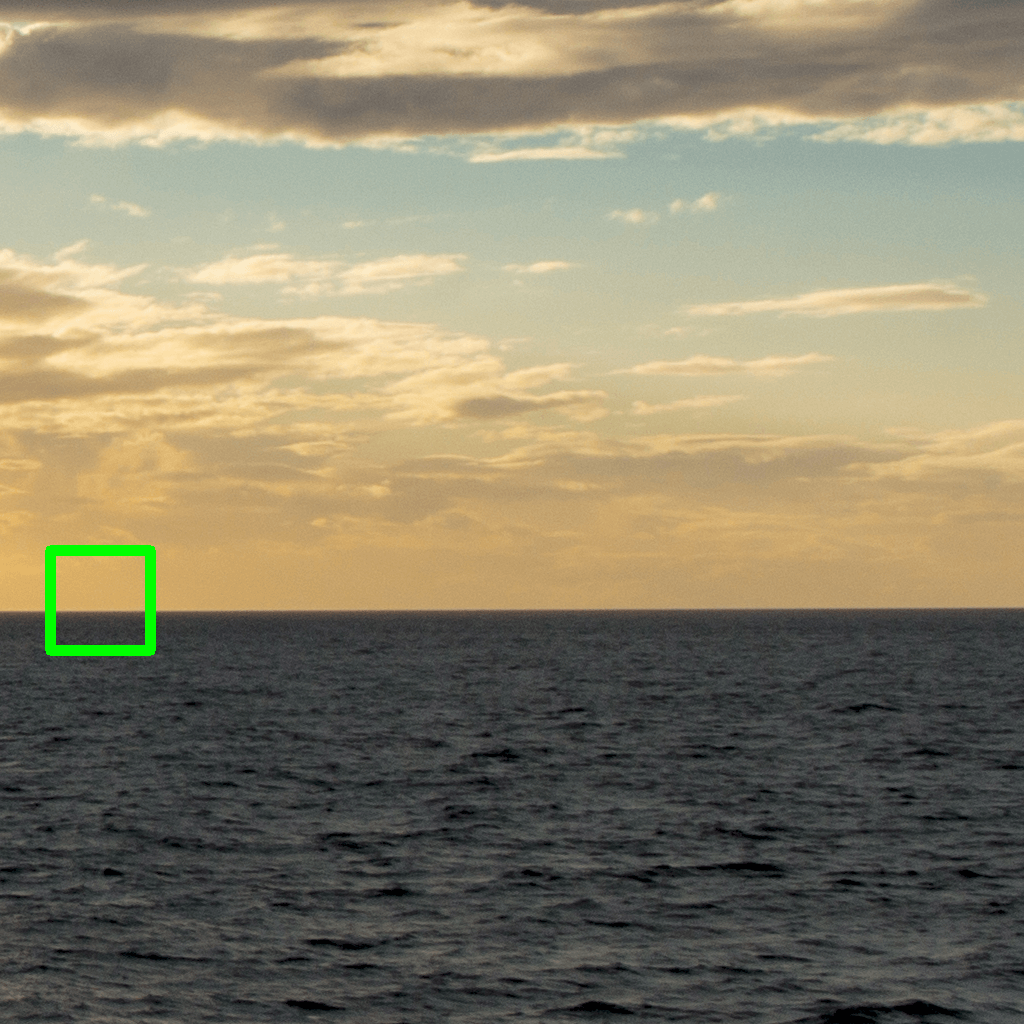}}
\subfigure{\centering\includegraphics[width=\wwsr, height=\hhsr]{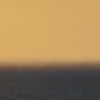}}
\subfigure{\centering\includegraphics[width=\wwsr, height=\hhsr]{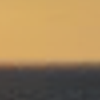}}
\subfigure{\centering\includegraphics[width=\wwsr, height=\hhsr]{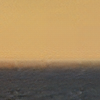}}
\subfigure{\centering\includegraphics[width=\wwsr, height=\hhsr]{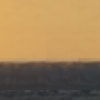}}
\par\bigskip\vspace*{-2em}
\subfigure{\centering\includegraphics[width=\wwsr, height=\hhsr]{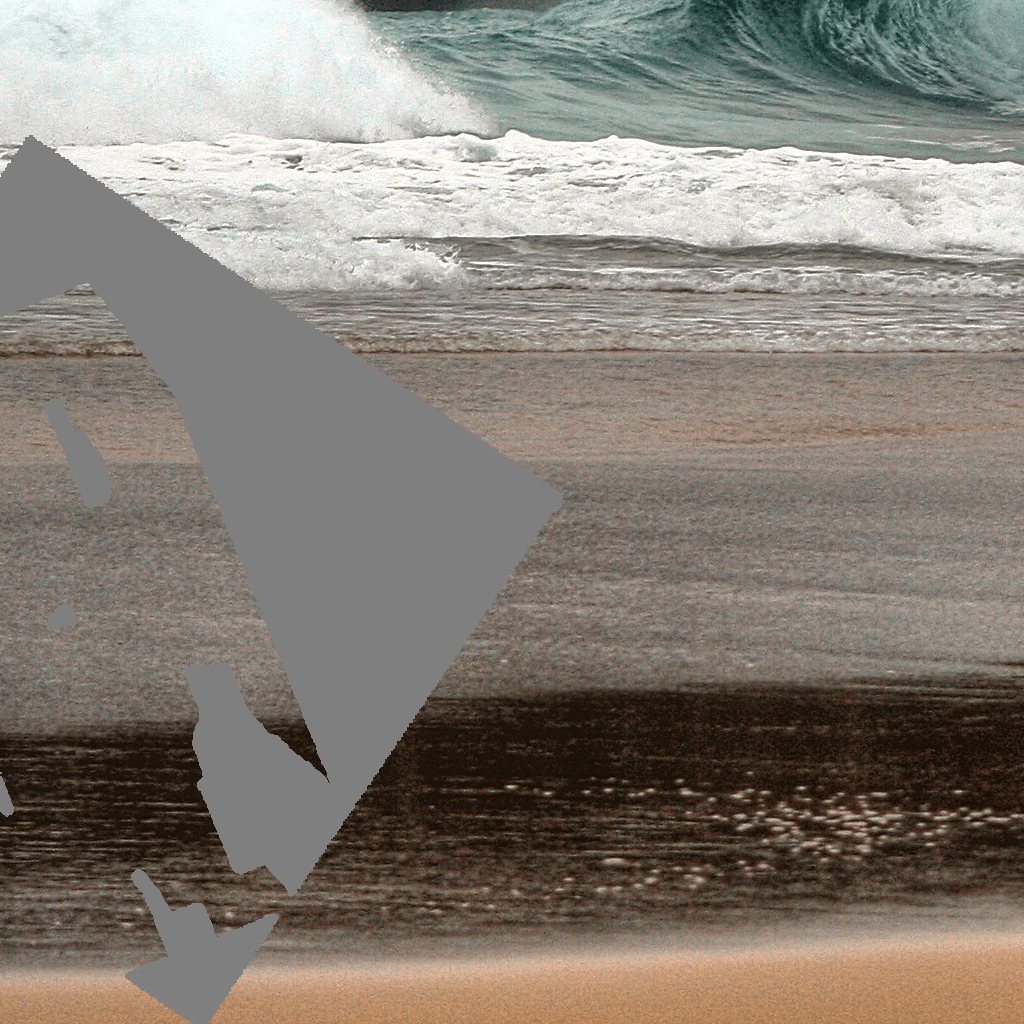}}
\subfigure{\centering\includegraphics[width=\wwsr, height=\hhsr]{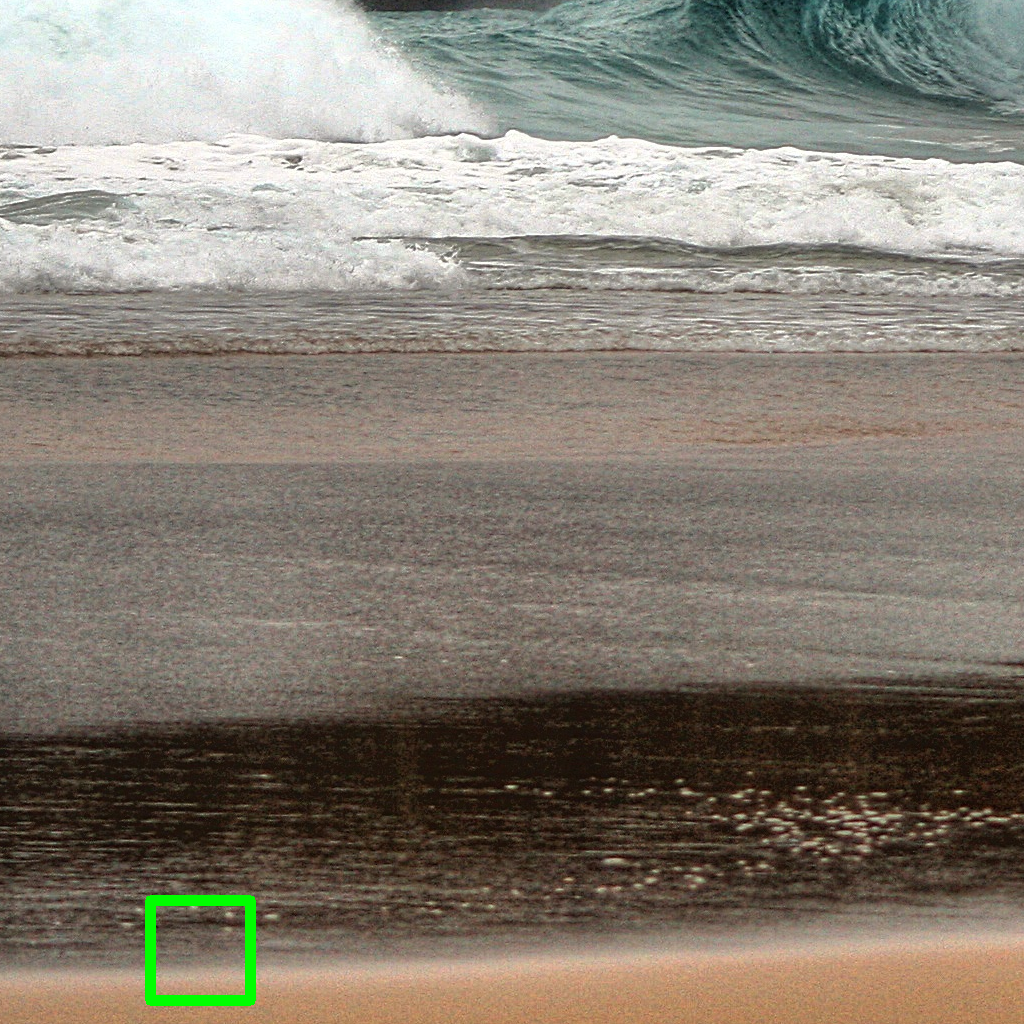}}
\subfigure{\centering\includegraphics[width=\wwsr, height=\hhsr]{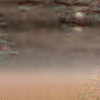}}
\subfigure{\centering\includegraphics[width=\wwsr, height=\hhsr]{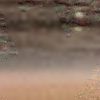}}
\subfigure{\centering\includegraphics[width=\wwsr, height=\hhsr]{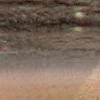}}
\subfigure{\centering\includegraphics[width=\wwsr, height=\hhsr]{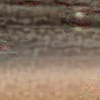}}
\par\bigskip\vspace*{-2em}
\addtocounter{subfigure}{-12}
\subfigure[\tiny{Input}]{\centering\includegraphics[width=\wwsr, height=\hhsr]{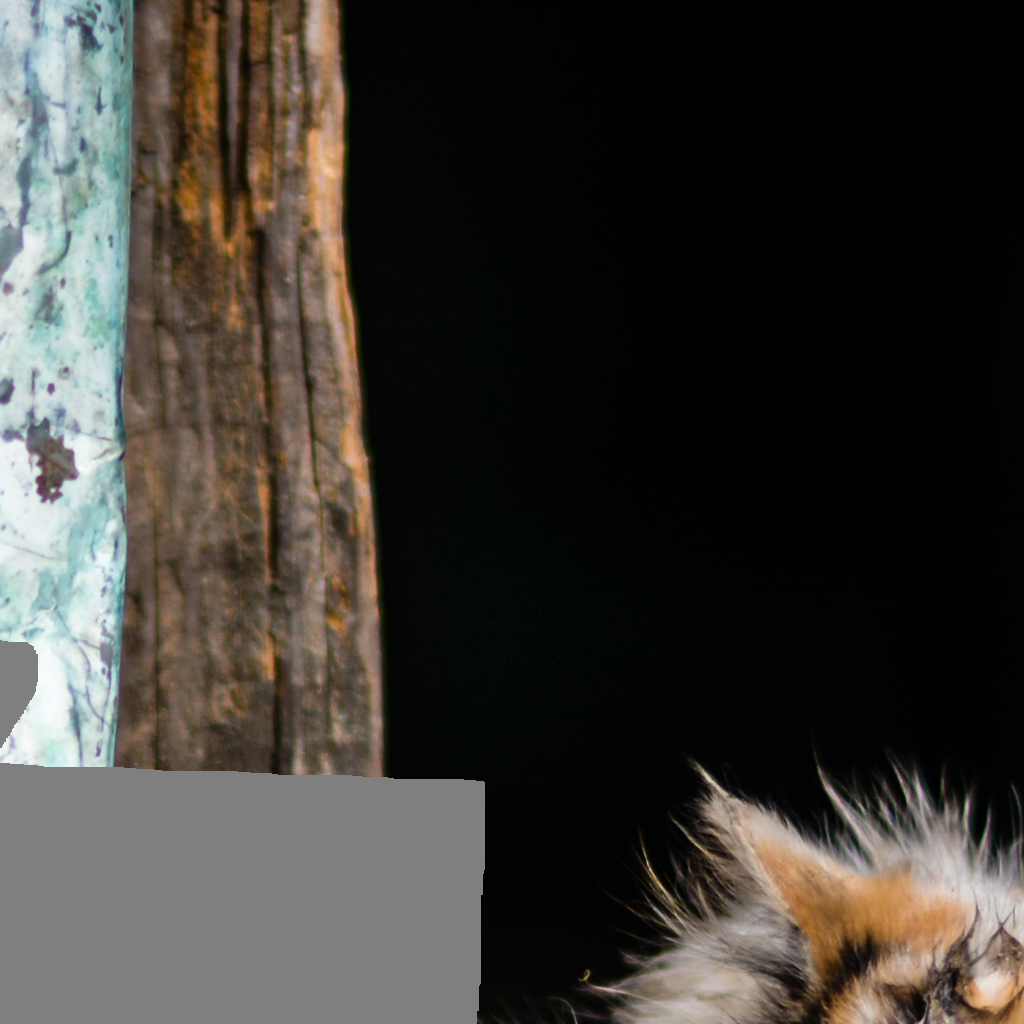}}
\subfigure[\tiny{GT}]{\centering\includegraphics[width=\wwsr, height=\hhsr]{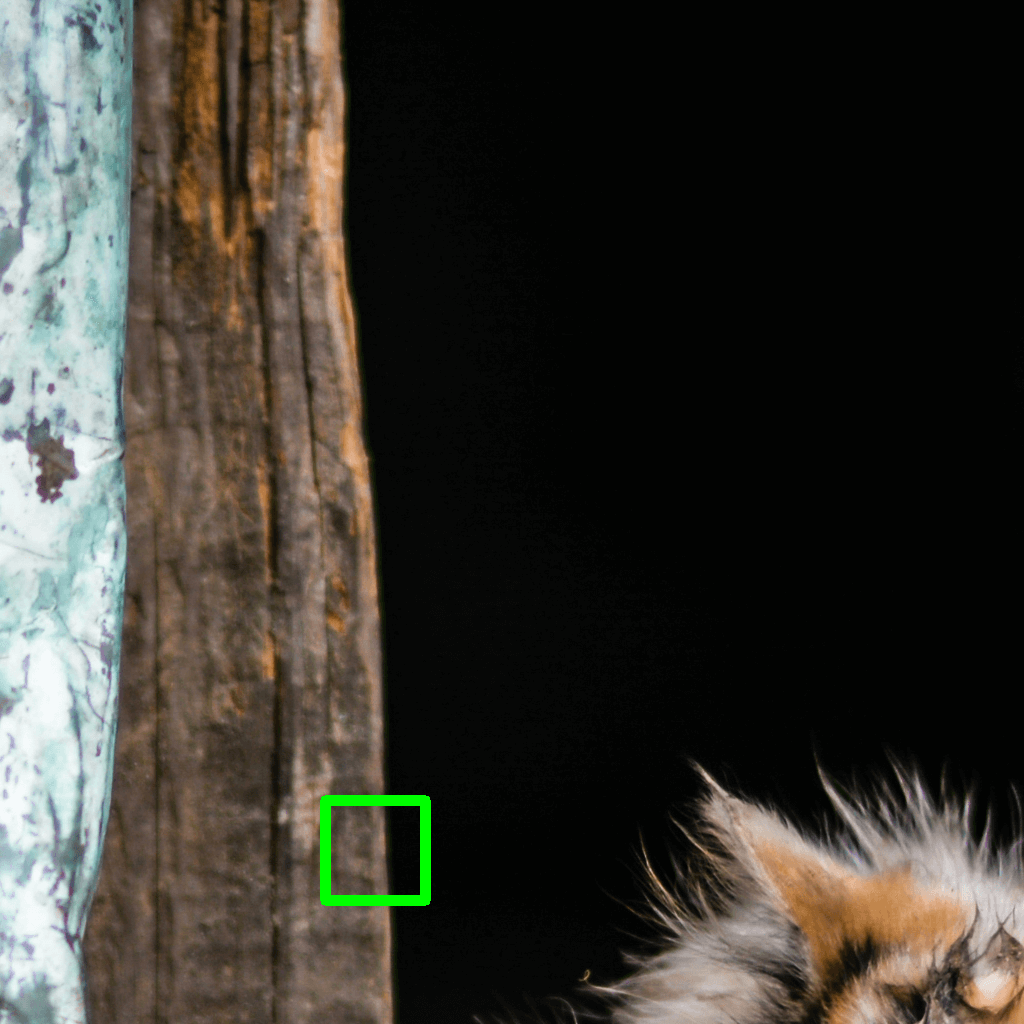}}
\subfigure[\tiny{Bilinear}]{\centering\includegraphics[width=\wwsr, height=\hhsr]{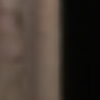}}
\subfigure[\tiny{Bicubic}]{\centering\includegraphics[width=\wwsr, height=\hhsr]{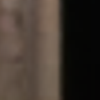}}
\subfigure[\tiny{SRGAN}]{\centering\includegraphics[width=\wwsr, height=\hhsr]{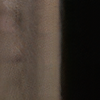}}
\subfigure[\tiny{CoordFill}]{\centering\includegraphics[width=\wwsr, height=\hhsr]{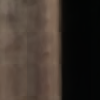}}

\vspace{-1em}
\caption{Qualitative comparisons of the proposed CoordFill and super-resolution methods on real-world high-resolution images.}
\label{fig:super-resolution}
\end{figure}

\newcommand\wwfailed{0.23\columnwidth}
\newcommand\hhfailed{0.23\columnwidth}
\begin{figure}[!h]
\centering  

\subfigure[\scriptsize{Input}]{\centering\includegraphics[width=\wwfailed, height=\hhfailed]{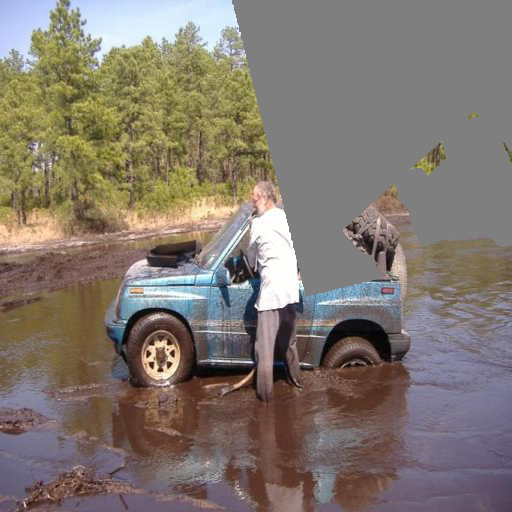}}
\subfigure[\scriptsize{GT}]{\centering\includegraphics[width=\wwfailed, height=\hhfailed]{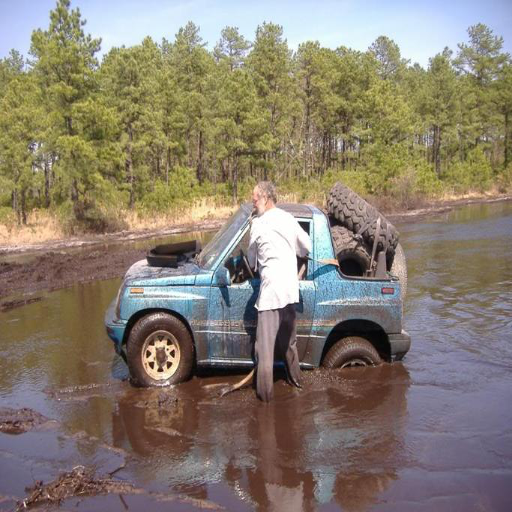}}
\subfigure[\scriptsize{LaMa}]{\centering\includegraphics[width=\wwfailed, height=\hhfailed]{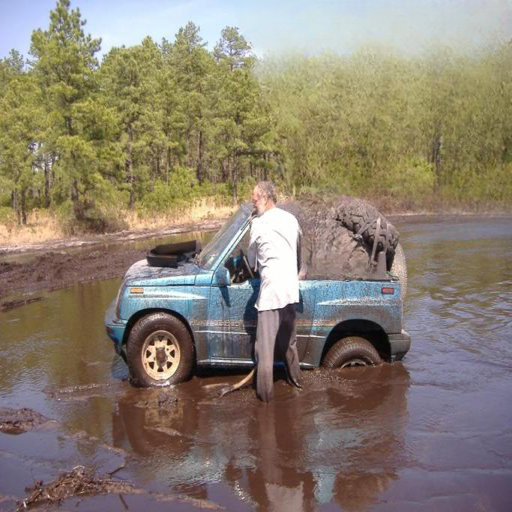}}
\subfigure[\scriptsize{CoordFill}]{\centering\includegraphics[width=\wwfailed, height=\hhfailed]{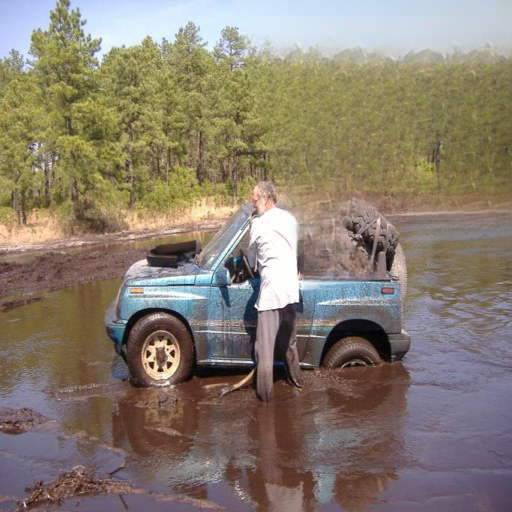}}

\vspace{-1em}
\caption{Failure case of the proposed CoordFill on Places2 512$\times$512 images. CoordFill gets dazzling artifacts when the background is complex. 
Although LaMa cannot achieve satisfactory results either, their artifacts are more natural.}
\label{fig:failed_case}
\end{figure}

\subsection{Limitation}
Although notable advantages are demonstrated by our method, there are still some limitations. 
To achieve stable and efficient performance, only a 256$\times$256 image will be used to infer the missing pixels, where the original high-frequency details are hard to preserve.
Besides, as shown in Figure~\ref{fig:failed_case}, in some cases, our method struggles to synthesize the photo-realistic textures when the background is complex, which is a common challenge of existing image inpainting methods. 
\section{Conclusion}
\label{sec:con}

We use the continuous coordinate representation for efficient high-resolution image inpainting. To this end, we propose CoordFill, a novel framework that contains the parameter generation network and the pixel-wise querying network following the meta-learning strategy. In detail, in the parameter generation network, we design a novel attention network based on FFC to produce the parameters in a spatial-adaptive fashion. In the pixel-wise querying network, we use the generated parameters from the parameter generation network and query the masked coordinate by re-sampling the coordinates in the frequency domain. The experiments show the efficiency of the proposed method on several datasets compared with several state-of-the-art methods.
\section{Acknowledgments}
\label{sec:acknowledgments}

This work was supported in part by the Science and Technology Development Fund, Macau SAR, underGrant 0034/2019/AMJ, Grant 0087/2020/A2, and Grant 0049/2021/A.


\begin{thebibliography}{45}
\providecommand{\natexlab}[1]{#1}

\bibitem[{Barnes et~al.(2009)Barnes, Shechtman, Finkelstein, and
  Goldman}]{patchmatch}
Barnes, C.; Shechtman, E.; Finkelstein, A.; and Goldman, D.~B. 2009.
\newblock {PatchMatch}: A Randomized Correspondence Algorithm for Structural
  Image Editing.
\newblock \emph{ACM Transactions on Graphics (Proc. SIGGRAPH)}, 28(3).

\bibitem[{Chen, Liu, and Wang(2021)}]{llif}
Chen, Y.; Liu, S.; and Wang, X. 2021.
\newblock Learning continuous image representation with local implicit image
  function.
\newblock In \emph{Proceedings of the IEEE/CVF Conference on Computer Vision
  and Pattern Recognition}, 8628--8638.

\bibitem[{Chi, Jiang, and Mu(2020)}]{ffc}
Chi, L.; Jiang, B.; and Mu, Y. 2020.
\newblock Fast fourier convolution.
\newblock \emph{Advances in Neural Information Processing Systems}, 33:
  4479--4488.

\bibitem[{Dong, Cao, and Fu(2022)}]{zits}
Dong, Q.; Cao, C.; and Fu, Y. 2022.
\newblock Incremental Transformer Structure Enhanced Image Inpainting with
  Masking Positional Encoding.
\newblock In \emph{Proceedings of the IEEE/CVF Conference on Computer Vision
  and Pattern Recognition}.

\bibitem[{Dosovitskiy et~al.(2020)Dosovitskiy, Beyer, Kolesnikov, Weissenborn,
  Zhai, Unterthiner, Dehghani, Minderer, Heigold, Gelly et~al.}]{vit}
Dosovitskiy, A.; Beyer, L.; Kolesnikov, A.; Weissenborn, D.; Zhai, X.;
  Unterthiner, T.; Dehghani, M.; Minderer, M.; Heigold, G.; Gelly, S.; et~al.
  2020.
\newblock An image is worth 16x16 words: Transformers for image recognition at
  scale.
\newblock \emph{arXiv preprint arXiv:2010.11929}.

\bibitem[{Goodfellow et~al.(2014)Goodfellow, Pouget-Abadie, Mirza, Xu,
  Warde-Farley, Ozair, Courville, and Bengio}]{gan}
Goodfellow, I.; Pouget-Abadie, J.; Mirza, M.; Xu, B.; Warde-Farley, D.; Ozair,
  S.; Courville, A.; and Bengio, Y. 2014.
\newblock Generative adversarial nets.
\newblock \emph{Advances in neural information processing systems}, 27.

\bibitem[{Howard et~al.(2017)Howard, Zhu, Chen, Kalenichenko, Wang, Weyand,
  Andreetto, and Adam}]{mobilenet}
Howard, A.~G.; Zhu, M.; Chen, B.; Kalenichenko, D.; Wang, W.; Weyand, T.;
  Andreetto, M.; and Adam, H. 2017.
\newblock Mobilenets: Efficient convolutional neural networks for mobile vision
  applications.
\newblock \emph{arXiv preprint arXiv:1704.04861}.

\bibitem[{Iandola et~al.(2016)Iandola, Han, Moskewicz, Ashraf, Dally, and
  Keutzer}]{squeezenet}
Iandola, F.~N.; Han, S.; Moskewicz, M.~W.; Ashraf, K.; Dally, W.~J.; and
  Keutzer, K. 2016.
\newblock SqueezeNet: AlexNet-level accuracy with 50x fewer parameters and< 0.5
  MB model size.
\newblock \emph{arXiv preprint arXiv:1602.07360}.

\bibitem[{Iizuka, Simo-Serra, and Ishikawa(2017)}]{global_and_local}
Iizuka, S.; Simo-Serra, E.; and Ishikawa, H. 2017.
\newblock Globally and locally consistent image completion.
\newblock \emph{ACM Transactions on Graphics (ToG)}, 36(4): 1--14.

\bibitem[{Isola et~al.(2017)Isola, Zhu, Zhou, and Efros}]{pix2pix}
Isola, P.; Zhu, J.-Y.; Zhou, T.; and Efros, A.~A. 2017.
\newblock Image-to-image translation with conditional adversarial networks.
\newblock In \emph{Proceedings of the IEEE conference on computer vision and
  pattern recognition}, 1125--1134.

\bibitem[{Jo, Yang, and Kim(2020)}]{jo2020investigating}
Jo, Y.; Yang, S.; and Kim, S.~J. 2020.
\newblock Investigating loss functions for extreme super-resolution.
\newblock In \emph{Proceedings of the IEEE/CVF conference on computer vision
  and pattern recognition workshops}, 424--425.

\bibitem[{Karras et~al.(2018)Karras, Aila, Laine, and Lehtinen}]{progan}
Karras, T.; Aila, T.; Laine, S.; and Lehtinen, J. 2018.
\newblock Progressive Growing of GANs for Improved Quality, Stability, and
  Variation.
\newblock In \emph{International Conference on Learning Representations}.

\bibitem[{Karras et~al.(2021)Karras, Aittala, Laine, H{\"a}rk{\"o}nen,
  Hellsten, Lehtinen, and Aila}]{styleganv3}
Karras, T.; Aittala, M.; Laine, S.; H{\"a}rk{\"o}nen, E.; Hellsten, J.;
  Lehtinen, J.; and Aila, T. 2021.
\newblock Alias-free generative adversarial networks.
\newblock \emph{Advances in Neural Information Processing Systems}, 34.

\bibitem[{Li et~al.(2020)Li, Lin, Ding, Liu, Zhu, and Han}]{gancompression}
Li, M.; Lin, J.; Ding, Y.; Liu, Z.; Zhu, J.-Y.; and Han, S. 2020.
\newblock GAN Compression: Efficient Architectures for Interactive Conditional
  GANs.
\newblock In \emph{Proceedings of the IEEE/CVF Conference on Computer Vision
  and Pattern Recognition}.

\bibitem[{Li et~al.(2022)Li, Lin, Zhou, Qi, Wang, and Jia}]{mat}
Li, W.; Lin, Z.; Zhou, K.; Qi, L.; Wang, Y.; and Jia, J. 2022.
\newblock MAT: Mask-Aware Transformer for Large Hole Image Inpainting.
\newblock In \emph{Proceedings of the IEEE/CVF Conference on Computer Vision
  and Pattern Recognition}.

\bibitem[{Liang, Cun, and Pun(2021)}]{s2crnet}
Liang, J.; Cun, X.; and Pun, C.-M. 2021.
\newblock Spatial-Separated Curve Rendering Network for Efficient and
  High-Resolution Image Harmonization.
\newblock \emph{arXiv preprint arXiv:2109.05750}.

\bibitem[{Liao et~al.(2021)Liao, Xiao, Wang, Lin, and Satoh}]{semantic_guide}
Liao, L.; Xiao, J.; Wang, Z.; Lin, C.-W.; and Satoh, S. 2021.
\newblock Image inpainting guided by coherence priors of semantics and
  textures.
\newblock In \emph{Proceedings of the IEEE/CVF Conference on Computer Vision
  and Pattern Recognition}, 6539--6548.

\bibitem[{Lin et~al.(2021)Lin, Lee, Cheng, Tulyakov, and Yang}]{infinitygan}
Lin, C.~H.; Lee, H.-Y.; Cheng, Y.-C.; Tulyakov, S.; and Yang, M.-H. 2021.
\newblock InfinityGAN: Towards Infinite-Pixel Image Synthesis.
\newblock \emph{arXiv preprint arXiv:2104.03963}.

\bibitem[{Liu et~al.(2018)Liu, Reda, Shih, Wang, Tao, and Catanzaro}]{pconv}
Liu, G.; Reda, F.~A.; Shih, K.~J.; Wang, T.-C.; Tao, A.; and Catanzaro, B.
  2018.
\newblock Image inpainting for irregular holes using partial convolutions.
\newblock In \emph{Proceedings of the European conference on computer vision
  (ECCV)}, 85--100.

\bibitem[{Liu et~al.(2020)Liu, Jiang, Song, Huang, and
  Yang}]{liu2020rethinking}
Liu, H.; Jiang, B.; Song, Y.; Huang, W.; and Yang, C. 2020.
\newblock Rethinking image inpainting via a mutual encoder-decoder with feature
  equalizations.
\newblock In \emph{European Conference on Computer Vision}, 725--741. Springer.

\bibitem[{Mildenhall et~al.(2020)Mildenhall, Srinivasan, Tancik, Barron,
  Ramamoorthi, and Ng}]{nerf}
Mildenhall, B.; Srinivasan, P.~P.; Tancik, M.; Barron, J.~T.; Ramamoorthi, R.;
  and Ng, R. 2020.
\newblock Nerf: Representing scenes as neural radiance fields for view
  synthesis.
\newblock In \emph{European conference on computer vision}, 405--421. Springer.

\bibitem[{Miyato et~al.(2018)Miyato, Kataoka, Koyama, and
  Yoshida}]{spectralnorm}
Miyato, T.; Kataoka, T.; Koyama, M.; and Yoshida, Y. 2018.
\newblock Spectral normalization for generative adversarial networks.
\newblock \emph{arXiv preprint arXiv:1802.05957}.

\bibitem[{Nazeri et~al.(2019)Nazeri, Ng, Joseph, Qureshi, and
  Ebrahimi}]{edgeconnect}
Nazeri, K.; Ng, E.; Joseph, T.; Qureshi, F.~Z.; and Ebrahimi, M. 2019.
\newblock Edgeconnect: Generative image inpainting with adversarial edge
  learning.
\newblock \emph{arXiv preprint arXiv:1901.00212}.

\bibitem[{Saito et~al.(2019)Saito, Huang, Natsume, Morishima, Kanazawa, and
  Li}]{pifu}
Saito, S.; Huang, Z.; Natsume, R.; Morishima, S.; Kanazawa, A.; and Li, H.
  2019.
\newblock Pifu: Pixel-aligned implicit function for high-resolution clothed
  human digitization.
\newblock In \emph{Proceedings of the IEEE/CVF International Conference on
  Computer Vision}, 2304--2314.

\bibitem[{Saito et~al.(2020)Saito, Simon, Saragih, and Joo}]{pifuhd}
Saito, S.; Simon, T.; Saragih, J.; and Joo, H. 2020.
\newblock Pifuhd: Multi-level pixel-aligned implicit function for
  high-resolution 3d human digitization.
\newblock In \emph{Proceedings of the IEEE/CVF Conference on Computer Vision
  and Pattern Recognition}, 84--93.

\bibitem[{Shaham et~al.(2021)Shaham, Gharbi, Zhang, Shechtman, and
  Michaeli}]{asapnet}
Shaham, T.~R.; Gharbi, M.; Zhang, R.; Shechtman, E.; and Michaeli, T. 2021.
\newblock Spatially-adaptive pixelwise networks for fast image translation.
\newblock In \emph{Proceedings of the IEEE/CVF Conference on Computer Vision
  and Pattern Recognition}, 14882--14891.

\bibitem[{Sitzmann et~al.(2020)Sitzmann, Martel, Bergman, Lindell, and
  Wetzstein}]{siren}
Sitzmann, V.; Martel, J.; Bergman, A.; Lindell, D.; and Wetzstein, G. 2020.
\newblock Implicit neural representations with periodic activation functions.
\newblock \emph{Advances in Neural Information Processing Systems}, 33:
  7462--7473.

\bibitem[{Suvorov et~al.(2022)Suvorov, Logacheva, Mashikhin, Remizova, Ashukha,
  Silvestrov, Kong, Goka, Park, and Lempitsky}]{lama}
Suvorov, R.; Logacheva, E.; Mashikhin, A.; Remizova, A.; Ashukha, A.;
  Silvestrov, A.; Kong, N.; Goka, H.; Park, K.; and Lempitsky, V. 2022.
\newblock Resolution-robust Large Mask Inpainting with Fourier Convolutions.
\newblock In \emph{Proceedings of the IEEE/CVF Winter Conference on
  Applications of Computer Vision}, 2149--2159.

\bibitem[{{Unsplash}(2021)}]{unsplash}
{Unsplash}. 2021.
\newblock Unsplash Lite Dataset 1.2.0.
\newblock [Online; accessed 4-March-2021].

\bibitem[{Vaswani et~al.(2017)Vaswani, Shazeer, Parmar, Uszkoreit, Jones,
  Gomez, Kaiser, and Polosukhin}]{transformer}
Vaswani, A.; Shazeer, N.; Parmar, N.; Uszkoreit, J.; Jones, L.; Gomez, A.~N.;
  Kaiser, {\L}.; and Polosukhin, I. 2017.
\newblock Attention is all you need.
\newblock \emph{Advances in neural information processing systems}, 30.

\bibitem[{Vo, Duong, and P{\'e}rez(2018)}]{structural_inpainting}
Vo, H.~V.; Duong, N.~Q.; and P{\'e}rez, P. 2018.
\newblock Structural inpainting.
\newblock In \emph{Proceedings of the 26th ACM international conference on
  Multimedia}, 1948--1956.

\bibitem[{Wan et~al.(2021)Wan, Zhang, Chen, and Liao}]{Wan_2021_ICCV}
Wan, Z.; Zhang, J.; Chen, D.; and Liao, J. 2021.
\newblock High-Fidelity Pluralistic Image Completion With Transformers.
\newblock In \emph{Proceedings of the IEEE/CVF International Conference on
  Computer Vision (ICCV)}, 4692--4701.

\bibitem[{Wang et~al.(2018)Wang, Liu, Zhu, Tao, Kautz, and
  Catanzaro}]{pix2pixhd}
Wang, T.-C.; Liu, M.-Y.; Zhu, J.-Y.; Tao, A.; Kautz, J.; and Catanzaro, B.
  2018.
\newblock High-Resolution Image Synthesis and Semantic Manipulation with
  Conditional GANs.
\newblock In \emph{Proceedings of the IEEE Conference on Computer Vision and
  Pattern Recognition}.

\bibitem[{Yi et~al.(2020)Yi, Tang, Azizi, Jang, and Xu}]{hifill}
Yi, Z.; Tang, Q.; Azizi, S.; Jang, D.; and Xu, Z. 2020.
\newblock Contextual residual aggregation for ultra high-resolution image
  inpainting.
\newblock In \emph{Proceedings of the IEEE/CVF Conference on Computer Vision
  and Pattern Recognition}, 7508--7517.

\bibitem[{Yu et~al.(2018)Yu, Lin, Yang, Shen, Lu, and Huang}]{deepfill}
Yu, J.; Lin, Z.; Yang, J.; Shen, X.; Lu, X.; and Huang, T.~S. 2018.
\newblock Generative image inpainting with contextual attention.
\newblock In \emph{Proceedings of the IEEE conference on computer vision and
  pattern recognition}, 5505--5514.

\bibitem[{Yu et~al.(2019)Yu, Lin, Yang, Shen, Lu, and Huang}]{deepfillv2}
Yu, J.; Lin, Z.; Yang, J.; Shen, X.; Lu, X.; and Huang, T.~S. 2019.
\newblock Free-form image inpainting with gated convolution.
\newblock In \emph{Proceedings of the IEEE/CVF International Conference on
  Computer Vision}, 4471--4480.

\bibitem[{Yu et~al.(2020)Yu, Guo, Jin, Wu, Chen, Li, Zhang, and Liu}]{rn}
Yu, T.; Guo, Z.; Jin, X.; Wu, S.; Chen, Z.; Li, W.; Zhang, Z.; and Liu, S.
  2020.
\newblock Region normalization for image inpainting.
\newblock In \emph{Proceedings of the AAAI Conference on Artificial
  Intelligence}, volume~34, 12733--12740.

\bibitem[{Zeng et~al.(2020{\natexlab{a}})Zeng, Cai, Li, Cao, and Zhang}]{3dlut}
Zeng, H.; Cai, J.; Li, L.; Cao, Z.; and Zhang, L. 2020{\natexlab{a}}.
\newblock Learning Image-adaptive 3D Lookup Tables for High Performance Photo
  Enhancement in Real-time.
\newblock \emph{IEEE Transactions on Pattern Analysis and Machine
  Intelligence}.

\bibitem[{Zeng et~al.(2022)Zeng, Fu, Chao, and Guo}]{aotgan}
Zeng, Y.; Fu, J.; Chao, H.; and Guo, B. 2022.
\newblock Aggregated contextual transformations for high-resolution image
  inpainting.
\newblock \emph{IEEE Transactions on Visualization and Computer Graphics}.

\bibitem[{Zeng et~al.(2021)Zeng, Lin, Lu, and Patel}]{crfill}
Zeng, Y.; Lin, Z.; Lu, H.; and Patel, V.~M. 2021.
\newblock Cr-fill: Generative image inpainting with auxiliary contextual
  reconstruction.
\newblock In \emph{Proceedings of the IEEE/CVF International Conference on
  Computer Vision}, 14164--14173.

\bibitem[{Zeng et~al.(2020{\natexlab{b}})Zeng, Lin, Yang, Zhang, Shechtman, and
  Lu}]{zeng2020high}
Zeng, Y.; Lin, Z.; Yang, J.; Zhang, J.; Shechtman, E.; and Lu, H.
  2020{\natexlab{b}}.
\newblock High-resolution image inpainting with iterative confidence feedback
  and guided upsampling.
\newblock In \emph{European conference on computer vision}, 1--17. Springer.

\bibitem[{Zhang et~al.(2018{\natexlab{a}})Zhang, Hu, Luo, Zuo, and
  Wang}]{semantic_inpainting}
Zhang, H.; Hu, Z.; Luo, C.; Zuo, W.; and Wang, M. 2018{\natexlab{a}}.
\newblock Semantic image inpainting with progressive generative networks.
\newblock In \emph{Proceedings of the 26th ACM international conference on
  Multimedia}, 1939--1947.

\bibitem[{Zhang et~al.(2018{\natexlab{b}})Zhang, Isola, Efros, Shechtman, and
  Wang}]{lpips}
Zhang, R.; Isola, P.; Efros, A.~A.; Shechtman, E.; and Wang, O.
  2018{\natexlab{b}}.
\newblock The unreasonable effectiveness of deep features as a perceptual
  metric.
\newblock In \emph{Proceedings of the IEEE conference on computer vision and
  pattern recognition}, 586--595.

\bibitem[{Zheng et~al.(2019)Zheng, Qiao, Cao, and Lau}]{distraction-aware}
Zheng, Q.; Qiao, X.; Cao, Y.; and Lau, R.~W. 2019.
\newblock Distraction-aware shadow detection.
\newblock In \emph{Proceedings of the IEEE/CVF Conference on Computer Vision
  and Pattern Recognition}, 5167--5176.

\bibitem[{Zhou et~al.(2017)Zhou, Lapedriza, Khosla, Oliva, and
  Torralba}]{place_dataset}
Zhou, B.; Lapedriza, A.; Khosla, A.; Oliva, A.; and Torralba, A. 2017.
\newblock Places: A 10 million image database for scene recognition.
\newblock \emph{IEEE transactions on pattern analysis and machine
  intelligence}, 40(6): 1452--1464.

\end{thebibliography}

\end{document}